\PassOptionsToPackage{table}{xcolor}
\documentclass{fairmeta}
\usepackage[utf8]{inputenc}
\usepackage{amsmath,amssymb}
\usepackage{tabularx,array,adjustbox,pifont,pdflscape,float}
\graphicspath{{Figures/}}
\colorlet{cvprblue}{metablue}
\definecolor{ourblue}{RGB}{231,234,251}
\definecolor{deltagray}{RGB}{243,243,243}

\title{VideoTED: Tree-Guided Evidence Distillation for Open-Vocabulary Action Recognition}
\renewcommand\authorformat[2][]{\mbox{\sffamily\bfseries #2$^{#1}$}}
\renewcommand\affiliationformat[2][]{\mbox{\small $^{#1}$#2}}
\newcommand\authornewrow[2][]{\addtolist[#1]{#2}{\authorlist}{\authorformat}{,\protect\newline}}
\newcommand\affiliationnewrow[2][]{\addtolist[#1]{#2}{\affiliationlist}{\affiliationformat}{,\protect\newline}}
\author[1,2,\dagger]{Weijie Zhu}
\author[1,\dagger]{Hanyu Fu}
\author[3]{Yuzhe Zhang}
\author[4]{Zhaoyan Pan}
\author[1]{Feiran Liu}
\authornewrow[1]{Xin Wei}
\author[1]{Hongbo Sun}
\author[1]{Zhiyu Lin}
\author[1]{Ye Yuan}
\author[1]{Yi Pan}
\authornewrow[1]{Tianyi Ding}
\author[1]{Junxian Yang}
\author[5]{Haocong He}
\author[1,*]{Han Fang}
\author[1,*]{Hao Sun}

\affiliation[1]{China Telecom Artificial Intelligence Technology (Beijing) Co., Ltd.}
\affiliationnewrow[2]{UCAS}
\affiliation[3]{PKU}
\affiliation[4]{ZJU}
\affiliation[5]{SJTU}

\contribution[\dagger]{Equal contribution}
\contribution[*]{Corresponding authors}

\metadata[Project page]{\href{https://0xiaofu0.github.io/Video-TED/}{\raisebox{-1.6pt}{\includegraphics[trim=268bp 257bp 288bp 155bp,clip,height=10.5pt]{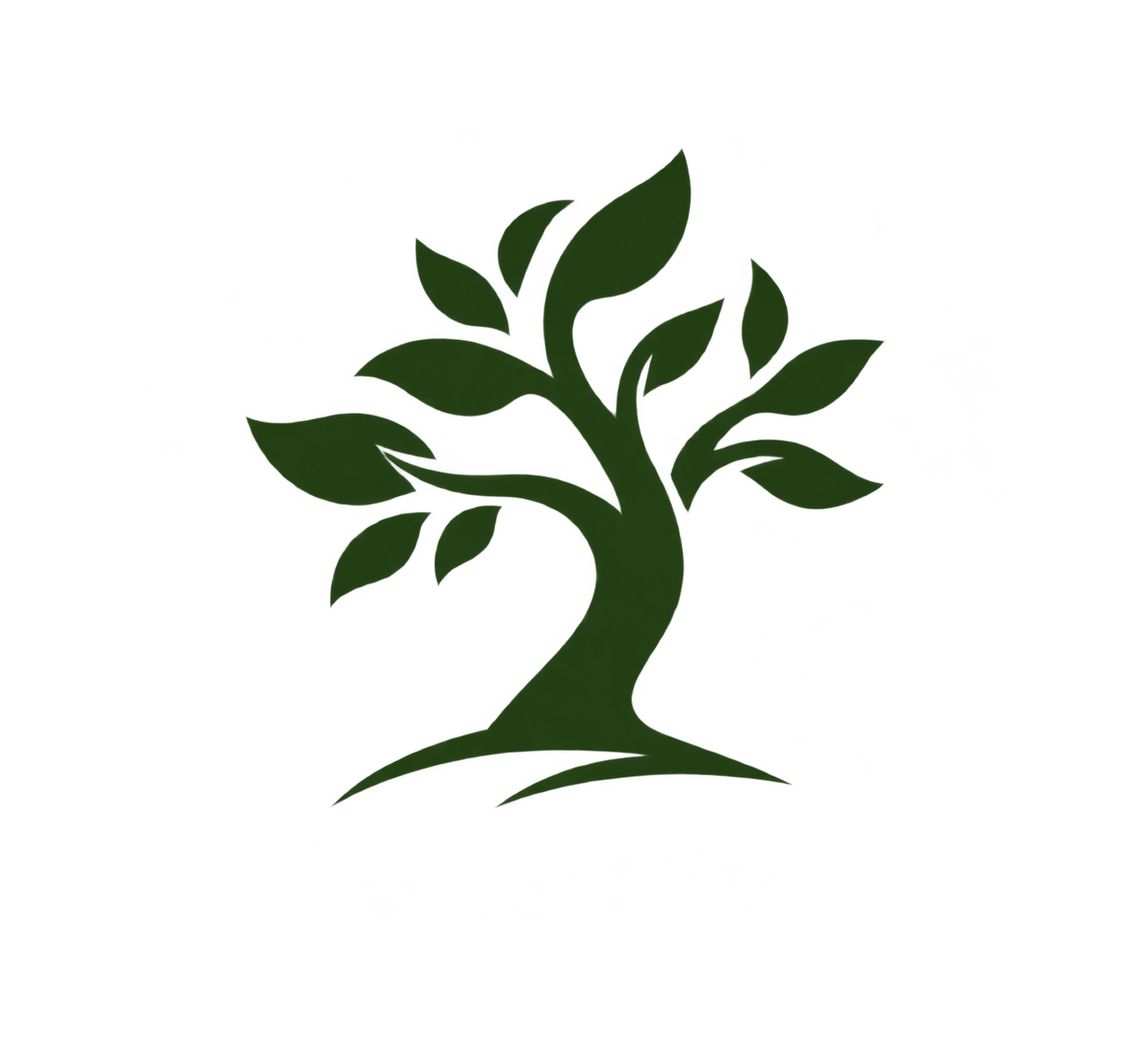}}\hspace{0.4em}\nolinkurl{https://0xiaofu0.github.io/Video-TED/}}}
\patchcmd{\mymaketitle}{\vskip 0.5cm}{\vskip 0.12cm}{}{%
  \ClassError{fairmeta}{Cannot set project-link spacing}{Check the title-box definition.}}
\patchcmd{\mymaketitle}{\setlength{\parskip}{0.5cm}}{\setlength{\parskip}{0.2cm}}{}{%
  \ClassError{fairmeta}{Cannot set title-box paragraph spacing}{Check the title-box definition.}}
\patchcmd{\mymaketitle}{\tcbset{bottom=0.35cm}}{\tcbset{bottom=0.15cm}}{}{%
  \ClassError{fairmeta}{Cannot set title-box bottom padding}{Check the title-box definition.}}
\hypersetup{pdftitle={VideoTED: Tree-Guided Evidence Distillation for Open-Vocabulary Action Recognition},
  pdfauthor={Weijie Zhu, Hanyu Fu, Yuzhe Zhang, Zhaoyan Pan, Feiran Liu, Xin Wei, Hongbo Sun, Zhiyu Lin, Ye Yuan, Yi Pan, Tianyi Ding, Junxian Yang, Haocong He, Han Fang, Hao Sun}}
\begin{document}
\begingroup
\setlength{\parskip}{4pt plus 1pt}
\setlength{\textfloatsep}{12pt plus 2pt minus 2pt}
\setlength{\floatsep}{10pt plus 2pt minus 2pt}
\setlength{\intextsep}{10pt plus 2pt minus 2pt}
\setlength{\abovedisplayskip}{7pt plus 2pt minus 2pt}
\setlength{\belowdisplayskip}{7pt plus 2pt minus 2pt}
\setlength{\abovedisplayshortskip}{3pt plus 1pt}
\setlength{\belowdisplayshortskip}{5pt plus 1pt minus 1pt}
\titlespacing*{\section}{0pt}{12pt plus 2pt minus 2pt}{6pt plus 1pt}
\titlespacing*{\subsection}{0pt}{9pt plus 2pt minus 2pt}{4pt plus 1pt}
\captionsetup{skip=4pt}
\abstract{Open-vocabulary action recognition (OVAR) requires grounding sub-actions and their compositional relations in visible evidence to recognize activity categories unseen during task-specific training. 
However, fine-grained descriptions often identify sub-actions without specifying the visual evidence for their relations, while outcome rewards provide limited guidance on whether these sub-actions and relations are correctly identified. 
In this work, we introduce \textbf{ActionTree-97k}, a dataset that organizes visible observations, sub-actions, and overall activities into Hierarchical Action Trees with explicit support relations. 
Building on this dataset, we propose \textbf{VideoTED}, a tree-guided evidence distillation framework for vision-language model post-training. 
Specifically, Tree-Conditioned Credit Modulation (TCM) uses a self-teacher conditioned on the action tree and activity label to modulate token-level update magnitudes on sampled responses while preserving reward-derived update directions. 
Reference Trajectory Learning (RTL) provides tree-derived reasoning traces and answers as independent learning targets excluded from group reward and advantage calculations, with more weight assigned when few sampled responses are correct. Across four OVAR benchmarks, VideoTED achieves absolute gains of up to \textbf{10.7\%} in base–novel harmonic mean accuracy over the base MLLM. On four general video-understanding benchmarks, average overall and action-related scores improve by \textbf{4.5\%} and \textbf{7.8\%}, respectively.}

\maketitle
\begin{figure}[!ht]
    \centering
    \includegraphics[width=0.92\linewidth]{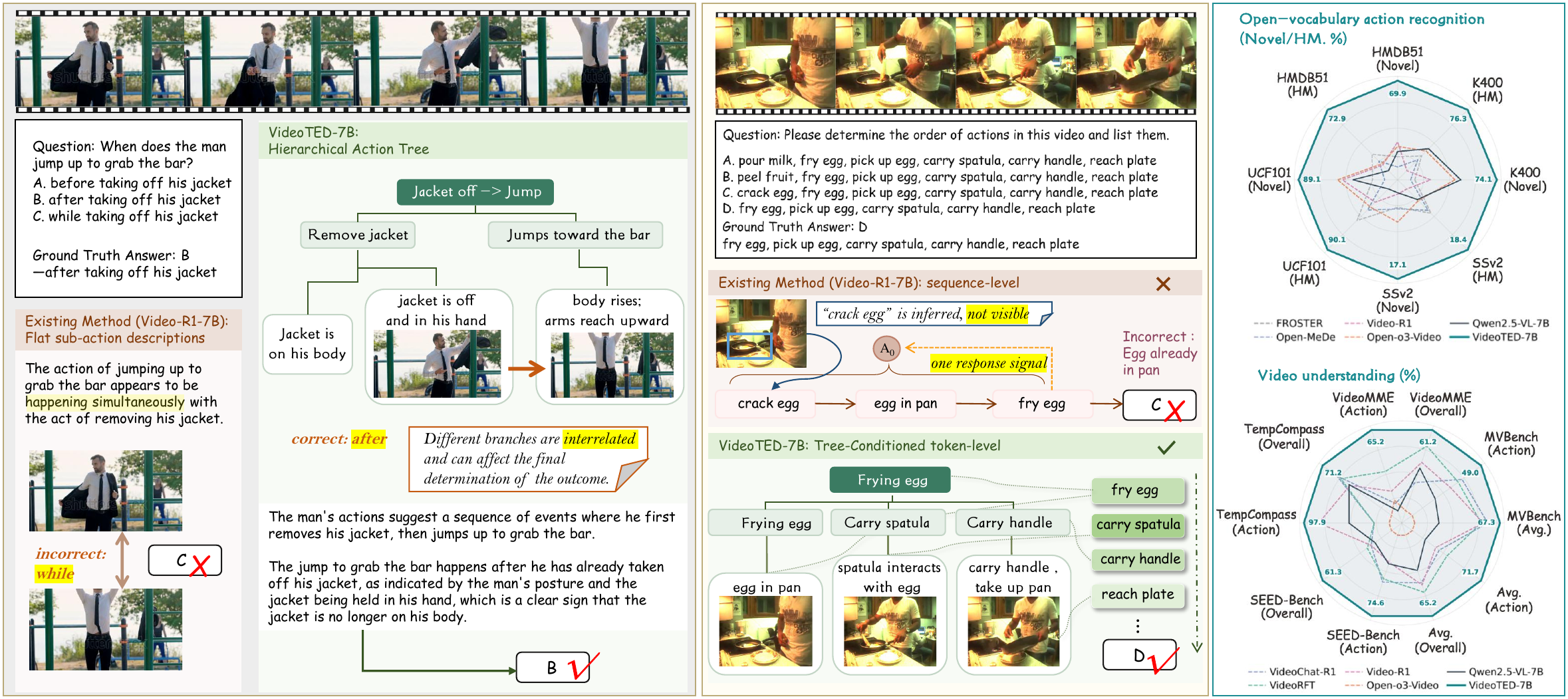}
\caption{Comparison of VideoTED with existing methods.
Left: VideoTED identifies jacket removal before jumping, while the baseline predicts simultaneous actions.
Middle: It grounds the frying-egg sequence in the sampled frames, avoiding an unsupported egg-cracking step.
Right: Performance comparison on open-vocabulary action recognition and general video-understanding benchmarks.}
    \label{fig:intro1}
\end{figure}
\section{Introduction}
\label{sec:intro}

Open-vocabulary action recognition (OVAR) aims to recognize activity categories unseen during task-specific training. Recent work has explored fine-grained action understanding for OVAR, using vision-language models (VLMs) to associate visual content with detailed action descriptions\cite{aghdam2025actalign}. These methods describe and decompose activities into sub-actions to capture local motion details\cite{liu2024sfar}. However, as shown in Fig.~\ref{fig:intro1}, recognizing an overall activity also requires understanding the temporal order and interactions among sub-actions. Fine-grained descriptions do not always explain these compositional relations, while outcome-based supervision offers limited guidance on the reasoning process. These limitations raise two challenges.



\ding{182}~\textbf{Fine-grained sub-actions require explicit compositional evidence.}
The same local motion can occur in different activities, and understanding its role requires considering surrounding sub-actions and object states. In Fig.~\ref{fig:intro1} (left), the baseline recognizes jacket removal and jumping but incorrectly predicts that they occur at the same time, whereas the sampled frames show jacket removal before jumping. Identifying individual sub-actions is therefore insufficient for overall activity recognition. A structured representation is needed to capture both sub-actions and their temporal and interaction relations.

\ding{183}~\textbf{Outcome rewards provide limited guidance on sub-actions and their relations.}
In Fig.~\ref{fig:intro1} (middle), the baseline includes an egg-cracking step that is not observed in the sampled frames, where the egg is already in the pan. A response-level outcome reward does not indicate which sub-actions or relations lack visual evidence. Moreover, when all responses in a sampled group fail, group-relative advantages become zero. Token-level reweighting therefore cannot provide an effective learning signal. This motivates evidence-guided token-level updates and independent reference supervision that remains effective even when all sampled responses fail.

To address these challenges, we introduce \textbf{ActionTree-97k} and \textbf{VideoTED} for open-vocabulary action recognition. ActionTree-97k organizes instance-level evidence into Hierarchical Action Trees, with observations at the leaves, sub-actions and their relations at the branches, and the activity at the root. This hierarchy grounds sub-actions and their composition in visible observations.

Building on this dataset, VideoTED integrates Tree-Conditioned Credit Modulation (TCM) and Reference Trajectory Learning (RTL). TCM conditions a self-teacher on the Hierarchical Action Tree and activity label, providing observations and compositional relations as context for scoring student-sampled tokens. Differences in log probabilities between the teacher and student modulate update magnitudes, while outcome rewards determine update directions. RTL converts the tree’s observations, sub-actions, and relations into a reference reasoning trace followed by the activity answer. The student learns this reference response as an independent target, with more weight assigned when few sampled responses are correct. Reference responses remain outside group reward and advantage calculations, allowing learning to continue even when all sampled responses are incorrect. VideoTED uses the observations and compositional relations encoded in action trees to improve recognition of activity categories unseen during task-specific training.
Our main contributions are:
\begin{itemize}
    \item We introduce ActionTree-97k, a training dataset for open-vocabulary action recognition that organizes visible observations, sub-actions, and overall activities into Hierarchical Action Trees with explicit support relations.
    \item We propose VideoTED, which combines Tree-Conditioned Credit Modulation (TCM) and Reference Trajectory Learning (RTL) to learn from hierarchical action evidence. TCM modulates token-level updates, while RTL provides independent reference supervision weighted by sampled success.
    \item Experiments on four OVAR and four general video-understanding benchmarks demonstrate consistent improvements over the base MLLM.
\end{itemize}


\section{Related Work}
\label{sec:related_work}

\subsection{Fine-grained and Compositional Action Understanding}

Understanding how fine-grained actions compose an overall activity is central to open-vocabulary action recognition (OVAR). To improve fine-grained discrimination, SFAR~\cite{liu2024sfar} uses a language model to generate atomic-action descriptions associated with activity classes. Task-Adapter++~\cite{cao2026taskadapterpp} generates ordered sub-action descriptions and aligns them with video stages. These methods capture local details and temporal order, but understanding an overall activity requires modeling how action components combine. RCORE~\cite{rcore2026} shows that compositional models may infer verbs from object co-occurrence instead of temporal evidence, and encourages sensitivity to temporal order. More directly related to OVAR, Video-STAR~\cite{yuan2025videostar} decomposes actions into contextual sub-motions for fine-grained reasoning, using tools and a hierarchical reward to emphasize relevant sub-motions. Yet these approaches do not  represent verified, instance-level support relations linking visible observations, sub-actions, and the overall activity. We therefore construct \textbf{ActionTree-97k} to provide traceable supervision through verified Hierarchical Action Trees.

\begin{figure}[t]
    \centering
    \begin{minipage}[b]{0.6\textwidth}
        \centering
        \includegraphics[width=\linewidth]{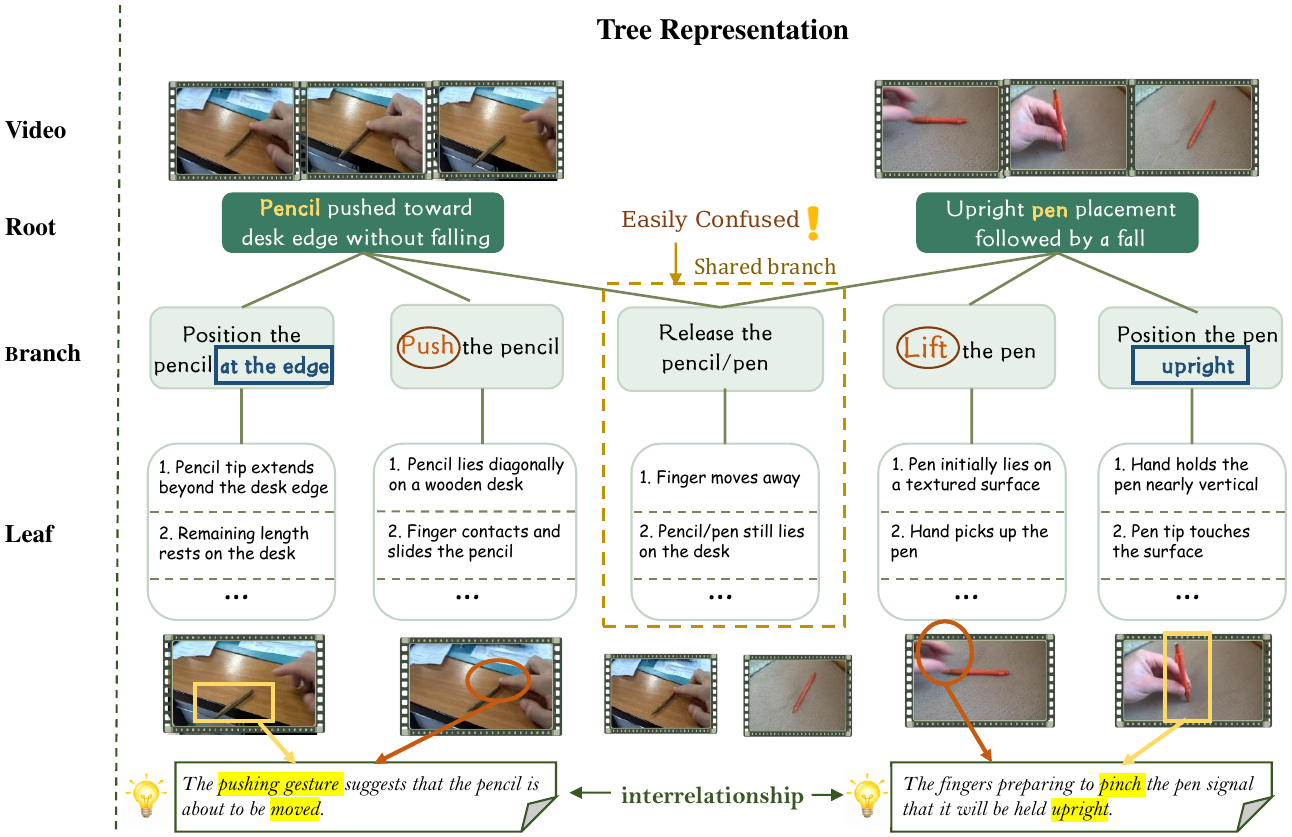}
        \par\smallskip
        (a)
    \end{minipage}\hfill 
    \quad
    \begin{minipage}[b]{0.35\textwidth}
        \centering
        \includegraphics[width=\linewidth]{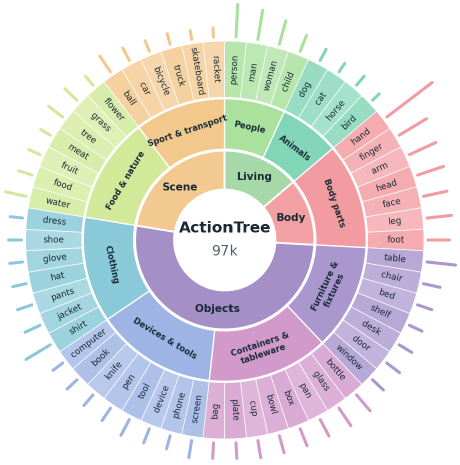}
        \par\smallskip
        (b)
    \end{minipage}
  \caption{
(a) \textbf{Hierarchical Action Tree generation.} Two easily confused activities share a release branch but differ in their action composition and supporting visible observations.
(b) \textbf{Concept coverage in ActionTree-97k.} The dataset covers scenes, living entities, body parts, and objects.
}
    \label{fig:action_tree_example}
\end{figure}

\subsection{Evidence-Guided Post-Training for VideoLLM}

Post-training improves video models by supervising or rewarding generated responses. VideoRFT~\cite{videorft2025} combines reasoning-trace supervision with a visual semantic-consistency reward, but response-level feedback cannot attribute intermediate judgments. Video-VER~\cite{luo2025videover} rewards visually grounded reasoning, while Open-o3-Video~\cite{meng2025openo3} supervises spatio-temporal evidence; neither identifies which parts of a sampled analysis need stronger updates. For token-level credit, Video-OPSD~\cite{wang2026videoopsd} builds a self-teacher from evidence frames, RLSD~\cite{rlsd2026} separates outcome-based update directions from teacher-modulated token magnitudes, and VISD~\cite{lin2026visd} applies this separation to structured video feedback. Yet these methods neither verify support relations from visible observations to sub-action composition nor provide updates when all sampled responses fail. OraRL~\cite{li2026orarl} adds oracle rollouts for failed cases, but mixing them into sampled groups distorts group reward and advantage statistics. VideoTED addresses both limitations through verified Hierarchical Action Trees: each conditions the self-teacher in TCM and supplies a success-weighted reference target outside group statistics in RTL. More details are provided in the supplementary material(\textcolor{cvprblue}{~\S A}).

\section{Hierarchical Action Evidence Dataset Construction}
\label{sec:dataset}

We construct ActionTree-97k to provide hierarchical action supervision. Each example contains a Hierarchical Action Tree, which is converted into a teacher context and a reference response for VideoTED.

\subsection{Action Tree Generation}
\label{sec:tree_generation}

A Hierarchical Action Tree has three levels. The root node represents the overall activity. Branch nodes represent action composition, including sub-actions and relations. Leaf nodes record visible observations supporting each branch. Uncertainty nodes mark details that cannot be confirmed from the video and are not used as evidence for the activity.

For video $V_j$, we represent a candidate tree as:
\begin{equation}
\begin{aligned}
T_{jd} &= \left( h_{jd}, \left\{(s_{jdk}, O_{jdk})\right\}_{k=1}^{K_j}, U_{jd} \right), \\
O_{jdk} &= \left\{ {o_{jdk1},\ldots,o_{jdkL_{jk}}}\right\},
\end{aligned}
\label{eq:action_tree}
\end{equation}
where $h_{jd}$ is the root node, $s_{jdk}$ is a branch node, $O_{jdk}$ contains its leaf nodes, and $U_{jd}$ contains the uncertainty nodes. 

Figure~\ref{fig:action_tree_example} (a) illustrates why these relations matter. Two pencil-handling activities can share the release of a pencil. Their trees distinguish the activities through the preceding actions, the pencil's position, and its resulting state. 


Qwen3.8-27B reads sampled frames from $V_j$ and generates $D=4$ candidate trees. During generation, it receives no ground-truth label, candidate answers, or existing reasoning trace, reducing the risk of tailoring evidence to a supplied answer. A parser validates each tree's schema before verification. (annotation and dataset details in Supp.\textcolor{cvprblue}{~\S B}).

\subsection{Verification and Training Annotations}
\label{sec:tree_verification}

We automatically verify candidate trees in a separate pass using the same annotation model with a different prompt. Given the video $V_j$, candidate tree $T_{jd}$, and a shuffled list of candidate activities $\widetilde{\mathcal{A}}_j$, the verifier produces:
\begin{equation}
(v_{jd}, c_{jd}^{\mathrm{hier}}, \hat{a}_{jd})
= F_{\phi}^{\mathrm{ver}}
\left(V_j, T_{jd}, \widetilde{\mathcal{A}}_j\right).
\label{eq:tree_verification}
\end{equation}
Here, $v_{jd}$ checks whether the observations are visible, $c_{jd}^{\mathrm{hier}}$ checks whether the nodes support their parents, and $\hat{a}_{jd}$ is the predicted activity.

Let $e_{jd}$ be the evidence-sufficiency flag produced during tree generation. We retain a candidate only if:
\begin{equation}
\mathrm{keep}_{jd}
=
\mathbf{1}\!\left[
e_{jd}=1
\land v_{jd}=1
\land c_{jd}^{\mathrm{hier}}=1
\land \hat{a}_{jd}=a_j
\right],
\label{eq:tree_filter}
\end{equation}
where $a_j$ is the ground-truth activity. If several candidates pass, we retain only the first valid candidate per video, discarding those without any passing candidates.


For each selected tree $T_j$, we construct two text annotations with the label $a_j$:
\begin{equation}
z_j=\operatorname{Context}(a_j,T_j),
\qquad
y_j^{*}=\operatorname{Response}(T_j,a_j).
\label{eq:tree_annotations}
\end{equation}

The teacher context $z_j$ comprises the activity label and hierarchical text formatted as an action tree. The reference response $y_j^{*}$ synthesizes the tree's compositional evidence and relations into an analysis, concluding with the activity label. Manual inspection confirms that 93.6\% of 1,000 randomly sampled trees are acceptable (Supp.\textcolor{cvprblue}{~\S E.1}).

\section{Method}
\label{sec:method}

\subsection{Overview}
\label{sec:overview}


VideoTED turns each verified action tree $T_j$ into two complementary learning paths, as shown in Figure~\ref{fig:method1}. Offline, $T_j$ and its label $a_j$ are formatted into a teacher context $z_j$ and a reference response $y_j^*$. Online, the student samples responses from the video and question alone and receives outcome rewards. \textbf{Tree-Conditioned Credit Modulation (TCM)} uses a $z_j$-conditioned self-teacher to modulate token-level update magnitudes on sampled responses, while \textbf{Reference Trajectory Learning (RTL)} learns from $y_j^*$ outside group statistics when successful samples are scarce. At inference, only the student model is used.

\begin{figure}[ht]                
    \centering
    \includegraphics[width=\linewidth]{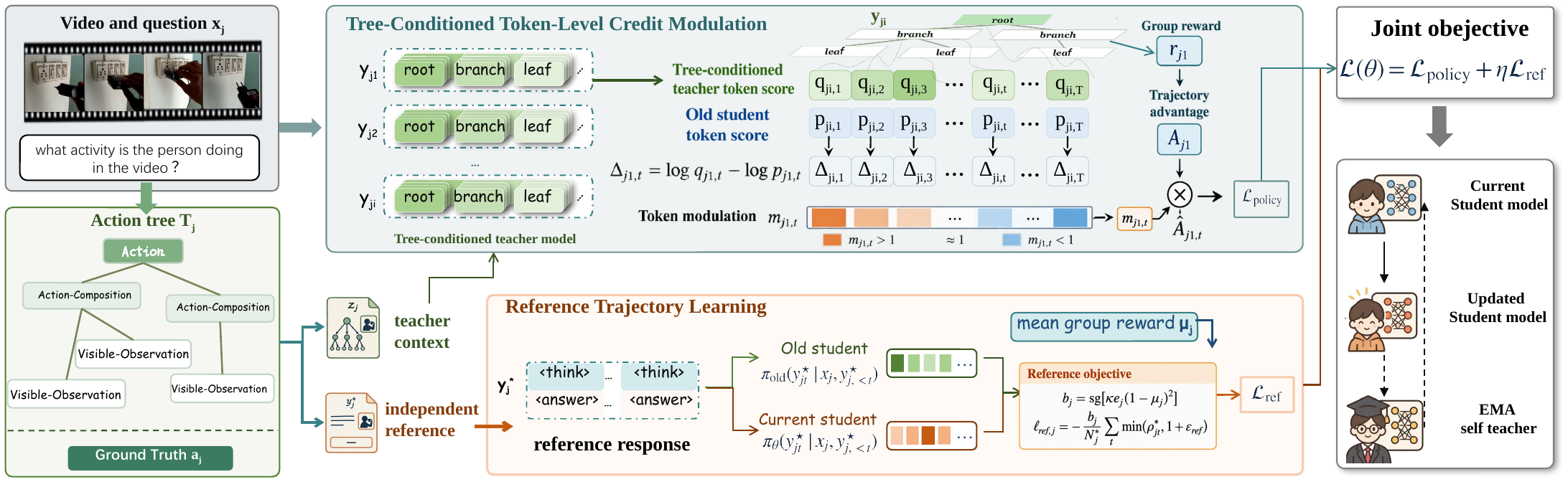}
    \caption{
\textbf{Overview of VideoTED. }A verified action tree provides a teacher context and an independent reference response. Tree-Conditioned Credit Modulation uses the teacher context to adjust token-level advantages on student samples, while Reference Trajectory Learning learns from the reference response with a weight determined by group success. The two losses jointly update the student, which in turn updates the EMA self-teacher.
}
    \label{fig:method1}
\end{figure}

\subsection{Tree-Conditioned Credit Modulation}
\label{sec:tcm}

If training relies only on whether the final activity prediction is correct, the observations and relations in the action tree cannot guide updates to individual tokens. To address this, we introduce tree-conditioned token-level credit modulation. The student first samples a response from the original video and question and receives an outcome reward. A self-teacher then scores the same generated tokens under the tree context. The probability difference between the teacher and old student modulates each token's update magnitude, while the outcome reward determines its direction.

\subsubsection{On-Policy Privileged Self-Teacher Guidance}
\label{sec:self_teacher}

We first assess task success by checking the student's final activity prediction. For example $j$, let $x_j$ be the video and question, and $a_j$ the ground-truth activity. Before each update, the old student policy samples $G$ responses as $y_{ji}\sim\pi_{\mathrm{old}}(\cdot\mid x_j)$, 
where$i=1,\ldots,G.$

Let $\operatorname{parse}(y_{ji})$ be the predicted activity and $\operatorname{valid}(y_{ji})$ indicate whether the analysis and answer are nonempty. We define the outcome reward and group mean as:
\begin{equation}
\begin{aligned}
    r_{ji}
&=
\mathbf{1}\!\left[\operatorname{parse}(y_{ji})=a_j\right]
\mathbf{1}\!\left[\operatorname{valid}(y_{ji})\right],
\\
\mu_j&=\frac{1}{G}\sum_{i=1}^{G}r_{ji},
\end{aligned}
\label{eq:outcome_reward}
\end{equation}
we compute a group-relative advantage for each sampled response:
\begin{equation}
\begin{aligned}
\sigma_j
&=
\sqrt{\frac{1}{G-1}
\sum_{i=1}^{G}(r_{ji}-\mu_j)^2},
\\
A_{ji}
&=
\operatorname{sg}\!\left(
\frac{r_{ji}-\mu_j}{\sigma_j+\epsilon}
\right),
\end{aligned}
\label{eq:group_advantage}
\end{equation}
where $\sigma_j$ is the sample standard deviation and $\operatorname{sg}(\cdot)$ stops gradients. Only student samples enter these statistics; reference responses do not. 

The advantage assigns one signal to an entire response. To refine credit within it, a self-teacher evaluates each sampled token under the teacher context $z_j$. For token $y_{jit}$, the old student and self-teacher score the same token under the same sampled prefix:
\begin{equation}
\begin{aligned}
p_{jit}
&=
\pi_{\mathrm{old}}(y_{jit}\mid x_j,y_{ji,<t}),\\
q_{jit}
&=
\pi_T(y_{jit}\mid x_j,z_j,y_{ji,<t}).
\end{aligned}
\label{eq:teacher_probabilities}
\end{equation}
The teacher remains frozen during the student update and receives an EMA update afterward. We use the log-probability difference
\begin{equation}
\Delta_{jit}=\log q_{jit}-\log p_{jit}
\label{eq:teacher_difference}
\end{equation}
to measure how the additional context changes the score of the sampled token.

\subsubsection{Action-Tree Advantage Reweighting}
\label{sec:advantage_reweighting}

The teacher difference $\Delta_{jit}$ reflects the joint effect of the activity label and action tree, but does not establish if the response is correct. We therefore use $A_{ji}$ to determine the update direction and $\Delta_{jit}$ to adjust its magnitude. Define
\begin{equation}
\begin{aligned}
w_{jit} &= \exp\big(\gamma_{jit}\,\text{sign}(A_{ji})\Delta_{jit}\big),\\
\lambda_k
&=0.5\max\!\left(1-\frac{k-1}{K_T},\,0\right),\\
m_{jit}
&=(1-\lambda_k)+\lambda_k
\operatorname{clip}\!\left(
w_{jit},1-\epsilon_w,1+\epsilon_w
\right),\\
\widehat{A}_{jit}
&=\operatorname{sg}\!\left(A_{ji}m_{jit}\right).
\end{aligned}
\label{eq:token_advantage}
\end{equation}

The positive factor $m_{jit}$ scales $A_{ji}$ without changing its direction; when $A_{ji}=0$, the update remains zero. Here, $\gamma_{jit} \in \{\gamma_{\text{root}}, \gamma_{\text{branch}}, \gamma_{\text{leaf}}\}$ modulates credit sensitivity across hierarchical tree levels rather than serving as a binary factual label.

We optimize the current student policy with a clipped PPO objective. For each sampled token, define

\begin{equation}
\rho_{jit}(\theta)
=
\frac{
\pi_\theta(y_{jit}\mid x_j,y_{ji,<t})
}{
p_{jit}
}.
\label{eq:policy_ratio}
\end{equation}
The token-level objective is
\begin{equation}
\begin{aligned}
s_{jit}(\theta)
&=
\min\!\bigl[
\rho_{jit}(\theta)\widehat{A}_{jit},
\\
&\qquad
\operatorname{clip}\!\left(
\rho_{jit}(\theta),
1-\epsilon_{\mathrm{pg}},
1+\epsilon_{\mathrm{pg}}
\right)\widehat{A}_{jit}
\bigr],
\end{aligned}
\label{eq:clipped_policy_objective}
\end{equation}
where $\epsilon_{\mathrm{pg}}$ limits the change from the old student policy.

\subsection{Reference Trajectory Learning}
\label{sec:rtl}

When group rewards are uniform, TCM yields zero advantages. RTL instead uses the tree-derived reference response $y_j^*$ as an independent learning target. The student learns this response from the original input $x_j$; the reference does not enter the sampled group's reward or advantage calculations.

Let $e_j\in\{0,1\}$ indicate whether the reference response is eligible for training. We set its weight according to the group's success rate:
\begin{equation}
b_j
=
\operatorname{sg}\!\left[
\kappa e_j(1-\mu_j)^2
\right].
\label{eq:reference_weight}
\end{equation}
Here, $\kappa$ sets the maximum reference weight. The reference weight is largest when no sample succeeds and falls to zero when all succeed.Unlike SFT that applies a static weight, this dynamic weighting provides on-demand guidance without interfering with successful policy exploration.

For reference token $y_{jt}^*$, both the current and old student policies use the original input and the same reference prefix. Their probability ratio is
\begin{equation}
\rho_{jt}^*(\theta)
=
\frac{\pi_\theta(y_{jt}^*\mid x_j,y_{j,<t}^*)}
{\pi_{\mathrm{old}}(y_{jt}^*\mid x_j,y_{j,<t}^*)}.
\label{eq:reference_ratio}
\end{equation}
With $N_j^*$ valid reference tokens, the reference loss is
\begin{equation}
\ell_j^{\mathrm{ref}}
=
-\frac{b_j}{N_j^*}
\sum_{t=1}^{N_j^*}
\min\!\left(
\rho_{jt}^*(\theta),1+\epsilon_{\mathrm{ref}}
\right).
\label{eq:reference_loss}
\end{equation}
This objective increases the relative probability of reference tokens until the ratio reaches its upper bound.

RTL uses the student policy directly, the tree-conditioned teacher does not score the reference response. The reference trajectory provides a complete learning target grounded in video evidence and a positive signal when the sampled group lacks correct responses.

\subsection{Joint Objective and Training Procedure}
\label{sec:joint_training}

We optimize the sampled responses and reference responses in the same student update. 

Let $\mathcal{B}$ be the set of sampled responses in a batch. For the $i$-th response sampled for video $j$, let $M_{jit}$ mark valid tokens and $N_{ji}=\sum_t M_{jit}$. The policy loss is
\begin{equation}
\mathcal{L}_{\mathrm{policy}}
=
-\frac{1}{|\mathcal{B}|}
\sum_{(j,i)\in\mathcal{B}}
\frac{1}{N_{ji}}
\sum_t M_{jit}s_{jit}(\theta).
\label{eq:policy_loss}
\end{equation}

\begin{table}[t]
\centering
\caption{Open-vocabulary action recognition (\%).
HM: harmonic mean of Base and Novel accuracy.
Best and second-best results are \textbf{bold} and \underline{underlined}.
$\dagger$: locally evaluated under our unified evaluation protocol using publicly released checkpoints (details in Supp.~\S D); other results are cited from original papers.
$\Delta$: percentage-point gains over Qwen2.5-VL-7B$^\dagger$, computed before rounding.}
\label{tab:ovar}
\setlength{\tabcolsep}{3.6pt}
\renewcommand{\arraystretch}{1.08}
\small

\newcommand{\ovpad}[1]{\phantom{0}#1}

\resizebox{\textwidth}{!}{%
\begin{tabular}{l*{12}{r}}
\toprule
\multirow{2}{*}{Method} & \multicolumn{3}{c}{K400} & \multicolumn{3}{c}{HMDB51} & \multicolumn{3}{c}{UCF101} & \multicolumn{3}{c}{SSv2} \\
\cmidrule(lr){2-4}\cmidrule(lr){5-7}\cmidrule(lr){8-10}\cmidrule(lr){11-13}
& Base & Novel & HM & Base & Novel & HM & Base & Novel & HM & Base & Novel & HM \\
\midrule
ActionCLIP~\cite{wang2021actionclip} & 61.0 & 46.2 & 52.6 & 69.1 & 37.3 & 48.5 & 90.1 & 58.1 & 70.7 & 13.3 & 10.1 & 11.5 \\
X-CLIP~\cite{ni2022xclip} & 74.1 & 56.4 & 64.0 & 69.4 & 45.5 & 55.0 & 89.9 & 58.9 & 71.2 & 8.5 & 6.6 & 7.4 \\
VPT~\cite{ju2022vpt} & 69.7 & 37.6 & 48.8 & 46.2 & 16.0 & 23.8 & 90.5 & 40.4 & 55.8 & 8.3 & 5.3 & 6.4 \\
ST-Adapter~\cite{pan2022stadapter} & 73.6 & 62.0 & 67.3 & 65.3 & 48.9 & 55.9 & 85.5 & 76.8 & 80.9 & 9.3 & 8.4 & 8.8 \\
ViFi-CLIP~\cite{vificlip2022} & 76.4 & 61.1 & 67.9 & 73.8 & 53.3 & 61.9 & 92.9 & 67.7 & 78.3 & 16.2 & 12.1 & 13.9 \\
FROSTER~\cite{huang2024froster} & \underline{77.8} & 64.3 & 70.4 & 74.1 & 58.0 & \underline{65.1} & \textbf{95.3} & 80.0 & \underline{87.0} & \underline{18.3} & 12.2 & \underline{14.6} \\
Open-MeDe~\cite{yu2025openmede} & 77.2 & 63.8 & 69.9 & 73.6 & 56.4 & 63.9 & \underline{94.9} & 78.5 & 85.9 & 17.1 & 12.3 & 14.3 \\
AP-CLIP~\cite{jia2023apclip} & 77.2 & 64.1 & 70.0 & \underline{74.6} & 55.9 & 63.9 & 94.8 & 77.0 & 84.8 & 16.3 & 12.9 & 14.4 \\

Video-R1$^\dagger$~\cite{feng2025videor1}
& \ovpad{70.9} & \ovpad{66.2} & \ovpad{68.5}
& \ovpad{66.4} & \ovpad{\underline{60.3}} & \ovpad{63.2}
& \ovpad{86.5} & \ovpad{83.7} & \ovpad{85.1}
& \ovpad{13.0} & \ovpad{11.3} & \ovpad{12.1} \\

Open-o3-Video$^\dagger$~\cite{meng2025openo3}
& 73.1 & \ovpad{69.1} & \ovpad{71.0}
& \ovpad{67.7} & 59.6 & \ovpad{63.4}
& \ovpad{86.0} & \ovpad{\underline{84.3}} & \ovpad{85.1}
& \ovpad{13.4} & \ovpad{\underline{13.2}} & \ovpad{13.3} \\

Video-STAR-7B$^\dagger$~\cite{yuan2025videostar}
& \ovpad{69.5} & \ovpad{64.7} & \ovpad{67.0}
& \ovpad{51.7} & \ovpad{43.1} & \ovpad{47.0}
& \ovpad{80.0} & \ovpad{74.3} & \ovpad{77.1}
& \ovpad{6.4} & \ovpad{4.9} & \ovpad{5.5} \\

\midrule
Qwen2.5-VL-7B$^\dagger$~\cite{bai2025qwen25vl}
& \ovpad{72.8} & \ovpad{\underline{69.8}} & \ovpad{\underline{71.3}}
& \ovpad{66.0} & \ovpad{58.8} & \ovpad{62.2}
& \ovpad{85.0} & \ovpad{82.4} & \ovpad{83.7}
& \ovpad{15.9} & \ovpad{11.2} & \ovpad{13.2} \\

\rowcolor{ourblue}\textbf{VideoTED-7B (ours)}
& \textbf{\ovpad{78.7}} & \textbf{\ovpad{74.1}} & \textbf{\ovpad{76.3}}
& \textbf{\ovpad{76.2}} & \textbf{\ovpad{69.9}} & \textbf{\ovpad{72.9}}
& \ovpad{91.1} & \textbf{\ovpad{89.1}} & \textbf{\ovpad{90.1}}
& \textbf{\ovpad{20.0}} & \textbf{\ovpad{17.1}} & \textbf{\ovpad{18.4}} \\

\rowcolor{deltagray}$\Delta$ vs. Qwen2.5-VL-7B$^\dagger$
& \ovpad{+5.8} & \ovpad{+4.3} & \ovpad{+5.0}
& \ovpad{+10.2} & \ovpad{+11.1} & \ovpad{+10.7}
& \ovpad{+6.1} & \ovpad{+6.7} & \ovpad{+6.4}
& \ovpad{+4.1} & \ovpad{+5.9} & \ovpad{+5.3} \\
\bottomrule
\end{tabular}}
\end{table}

Each response is averaged over its valid tokens before the batch average, giving responses of different lengths equal weight.

Let $\mathcal{J}$ be the videos in the batch. The reference loss is
\begin{equation}
\mathcal{L}_{\mathrm{ref}}
=
\frac{1}{|\mathcal{J}|}
\sum_{j\in\mathcal{J}}\ell_j^{\mathrm{ref}}.
\label{eq:batch_reference_loss}
\end{equation}

We update the student with
\begin{equation}
\mathcal{L}
=
\mathcal{L}_{\mathrm{policy}}
+\eta\mathcal{L}_{\mathrm{ref}},
\label{eq:joint_objective}
\end{equation}

where $\eta$ balances the two losses. Reference responses are excluded from the sampled groups used to compute rewards and advantages. For the first $K_T$ steps, the frozen self-teacher scores sampled tokens using $z_j$ and receives an EMA update after each student update. Thereafter, the policy loss uses the unmodulated group advantage, while reference learning continues. At inference, only the student uses the video and question for analysis and activity prediction.

\section{Experiments}

\subsection{Experimental setup}

\noindent\textbf{Implementation details.}
We post-train Qwen2.5-VL (3B/7B)~\cite{bai2025qwen25vl} and Qwen3-VL (4B/8B)~\cite{bai2025qwen3vl} on ActionTree-97k, with Qwen2.5-VL-7B as the main backbone. 
Main experiments use 32 NVIDIA A800 (80GB) GPUs for one epoch.
We use AdamW with a learning rate of $10^{-6}$ and cosine decay.
vLLM samples four responses per prompt in batches of 32 videos.
Videos use up to 16 frames at 2 FPS; responses are capped at 768 tokens. (full training settings and hyperparameters in Supp.\textcolor{cvprblue}{~\S C}).

\noindent\textbf{Benchmarks.} We evaluate OVAR on Kinetics-400~\cite{kay2017kinetics}, HMDB51~\cite{kuehne2011hmdb}, UCF101~\cite{soomro2012ucf101}, and Something-Something V2~\cite{goyal2017something}, reporting accuracy on the base and novel partitions and their harmonic mean (HM). We further assess general video understanding on MVBench~\cite{li2023mvbench}, Video-MME~\cite{fu2024videomme} without subtitles, TempCompass~\cite{liu2024tempcompass}-MCQ, and the video subset of SEED-Bench~\cite{li2023seedbench}. For these benchmarks, we report overall and action-related scores, together with their respective unweighted averages across the four benchmarks. (detailed evaluation protocols in Supp.\textcolor{cvprblue}{~\S D}).

\subsection{Main Results}

\begin{table}[t]
\centering
\caption{General video understanding (\%).
Avg.\ denotes the unweighted mean across four benchmarks.
\textbf{Bold} and \underline{underline} indicate the best and second-best results, respectively.
$\dagger$: locally evaluated under our unified evaluation protocol using released checkpoints; LLaVA-OV-7B results are cited from the original paper.
$\Delta$ denotes percentage-point gains over Qwen2.5-VL-7B$^\dagger$, computed before rounding.
Dashes indicate unavailable results or incomplete averages.}
\label{tab:video}

\small
\setlength{\tabcolsep}{1.5pt}
\renewcommand{\arraystretch}{1.12}

\begin{tabularx}{\textwidth}{l*{10}{>{\centering\arraybackslash}X}}
\toprule
\multirow{2}{*}{\textbf{Model}}
& \multicolumn{2}{c}{\textbf{MVBench}}
& \multicolumn{2}{c}{\textbf{Video-MME}}
& \multicolumn{2}{c}{\textbf{TempCompass}}
& \multicolumn{2}{c}{\textbf{SEED-Bench}}
& \multicolumn{2}{c}{\textbf{Avg.}} \\
\cmidrule(lr){2-3}
\cmidrule(lr){4-5}
\cmidrule(lr){6-7}
\cmidrule(lr){8-9}
\cmidrule(lr){10-11}
& Avg. & Action
& Overall & Action
& Overall & Action
& Overall & Action
& Overall & Action \\
\midrule
LLaVA-OV-7B~\cite{li2024llavaonevision}
& 56.7 & --
& 58.2 & --
& 67.2 & --
& -- & --
& -- & -- \\
TW-GRPO$^\dagger$~\cite{dang2025twgrpo}
& 65.6 & \underline{47.5}
& 56.6 & 54.3
& 68.9 & 97.3
& \underline{57.9} & 66.5
& 62.2 & 66.4 \\
VideoChat-R1$^\dagger$~\cite{li2025videochatr1}
& \underline{66.0} & 46.5
& 54.7 & 52.1
& \underline{69.8} & 96.5
& 57.5 & \underline{68.1}
& 62.0 & 65.8 \\
VideoRFT$^\dagger$~\cite{videorft2025}
& 65.3 & 46.0
& \underline{59.2} & \underline{57.8}
& 69.4 & 96.5
& 57.8 & 67.5
& \underline{62.9} & \underline{66.9} \\
Video-STAR-7B$^\dagger$~\cite{yuan2025videostar}
& 54.2 & 40.0
& 49.6 & 45.1
& 64.8 & 96.2
& 50.6 & 53.0
& 54.8 & 58.5 \\
Video-R1$^\dagger$~\cite{feng2025videor1}
& 65.5 & 45.0
& 57.2 & 54.3
& 68.5 & \underline{97.6}
& 57.5 & 64.3
& 62.2 & 65.3 \\
Open-o3-Video$^\dagger$~\cite{meng2025openo3}
& 58.2 & 40.0
& 51.2 & 52.7
& 64.4 & 96.2
& 55.9 & 59.1
& 57.4 & 62.0 \\
\midrule
Qwen2.5-VL-7B$^\dagger$~\cite{bai2025qwen25vl}
& 60.8 & 43.0
& 56.5 & 50.8
& 68.6 & 97.0
& 57.2 & 64.6
& 60.8 & 63.8 \\
\rowcolor{ourblue}
\textbf{VideoTED-7B (ours)}
& \textbf{67.3} & \textbf{49.0}
& \textbf{61.2} & \textbf{65.2}
& \textbf{71.2} & \textbf{97.9}
& \textbf{61.3} & \textbf{74.6}
& \textbf{65.2} & \textbf{71.7} \\
\rowcolor{deltagray}
$\Delta$ vs. Qwen2.5-VL-7B$^\dagger$
& +6.6 & +6.0
& +4.7 & +14.4
& +2.6 & +0.9
& +4.1 & +10.1
& +4.5 & +7.8 \\
\bottomrule
\end{tabularx}
\end{table}

\newcommand{\abldrop}[2]{%
    #1\textsuperscript{\textcolor{gray}{(-#2)}}%
}


\noindent\textbf{Open-vocabulary recognition across diverse activities.}
We first evaluate whether VideoTED improves novel-category recognition while preserving base-category performance under a unified, frame-only protocol (Supp.~\S D).
Table~\ref{tab:ovar} shows gains on both partitions over Qwen2.5-VL-7B across K400, HMDB51, and UCF101, with the highest novel accuracy and HM among the compared methods on each benchmark.
For example, HM increases from 62.2\% to 72.9\% on HMDB51.
The simultaneous improvements on both partitions indicate that stronger novel-category recognition does not come at the expense of base-category performance in these evaluations.

\noindent\textbf{Fine-grained action recognition on SSv2.}
We further assess fine-grained recognition on SSv2, which focuses on object interactions.
In Table~\ref{tab:ovar}, VideoTED achieves the highest base and novel accuracies, reaching an HM of 18.4\%, compared with 13.2\% for Qwen2.5-VL-7B and 14.6\% for FROSTER.
Together with the preceding results, these improvements support VideoTED's effectiveness across different levels of action granularity, covering both broader activity categories and fine-grained interactions.

\noindent\textbf{General video understanding.}
We also examine whether action-focused post-training benefits broader video understanding.
Table~\ref{tab:video} shows that VideoTED leads all eight benchmark-specific metrics, achieving average overall and action-related scores of 65.2\% and 71.7\%, respectively.
For example, the action-related score on SEED-Bench increases from 64.6\% to 74.6\%.
The joint improvements in overall and action-related performance show that VideoTED strengthens specialized action recognition while also improving broader video understanding.
\providecommand{\abldrop}[2]{%
    #1\textsuperscript{\textcolor{gray}{$-#2$}}%
}

\begin{table}[t]
    \centering
    \caption{Ablation of TCM and RTL with a reference SFT baseline.
    The TCM/RTL variants retain GRPO.
    Gray superscripts denote score drops from the full model.}
    \label{tab:ablation_tcm_rtl}

    \begin{adjustbox}{max width=\linewidth}
        \begin{tabular}{ccccc}
            \toprule
            \textbf{TCM} & \textbf{RTL}
            & \textbf{General Avg.} $\uparrow$
            & \textbf{Action Avg.} $\uparrow$
            & \textbf{SSv2 HM} $\uparrow$ \\
            \midrule
            \multicolumn{2}{c}{Reference SFT only}
            & \abldrop{61.6}{3.6}
            & \abldrop{66.4}{5.3}
            & \abldrop{12.3}{6.1} \\
            \midrule
            \ding{55} & \ding{55}
            & \abldrop{62.4}{2.8}
            & \abldrop{68.5}{3.2}
            & \abldrop{15.7}{2.7} \\
            \ding{51} & \ding{55}
            & \abldrop{62.8}{2.4}
            & \abldrop{68.4}{3.3}
            & \abldrop{17.2}{1.2} \\
            \ding{55} & \ding{51}
            & \abldrop{62.8}{2.4}
            & \abldrop{69.2}{2.5}
            & \abldrop{18.3}{0.1} \\
            \specialrule{\lightrulewidth}{\aboverulesep}{0pt}
            \rowcolor{ourblue}
            \ding{51} & \ding{51}
            & \textbf{65.2}
            & \textbf{71.7}
            & \textbf{18.4} \\
            \specialrule{\heavyrulewidth}{0pt}{0pt}
        \end{tabular}
    \end{adjustbox}
\end{table}


\providecommand{\abldrop}[2]{%
    #1\textsuperscript{\textcolor{gray}{(-#2)}}%
}

\subsection{Ablation Studies}
We assess the complementarity of TCM and RTL in Table~\ref{tab:ablation_tcm_rtl} and their key design choices in Table~\ref{tab:component_ablation}.
General Avg.\ and Action Avg.\ denote the unweighted means of the overall and action-related scores, respectively, across the four video-understanding benchmarks in Table~\ref{tab:video}.



\noindent\textbf{Effect of TCM and RTL.}
Supervised fine-tuning (SFT) on reference analyses and answers achieves 61.6\% General Avg.\ and 12.3\% SSv2 HM, below the full model's 65.2\% and 18.4\%, showing that imitation alone is insufficient.
Adding TCM to GRPO~\cite{shao2024grpo} raises SSv2 HM from 15.7\% to 17.2\%, with little change in Action Avg.
RTL alone reaches 18.3\% SSv2 HM, nearly matching the full model.
Combining both yields the highest General and Action averages, supporting their complementarity.

\noindent\textbf{Effect of tree structure and teacher guidance.}
\textit{w/o Tree Structure} flattens teacher context and reference analyses, preserving evidence; \textit{w/o Tree Context} removes tree evidence from the teacher but retains the activity label.
Action Avg.\ falls from 71.7\% to 64.5\% and 63.6\%, respectively, highlighting the value of hierarchy and teacher conditioning.
Freezing the initial teacher (\textit{w/o EMA}) lowers SSv2 HM to 11.3\%, supporting teacher updates.

\noindent\textbf{Effect of reference supervision.}
Answer-only targets (\textit{w/o Ref. Analysis}) reduce Action Avg.\ from 71.7\% to 65.3\%, demonstrating the value of analyses beyond final answers.
Uniform reference weights (\textit{w/o Ref. Weighting}) lower it to 68.6\%, supporting weighting by group success rate.

\noindent\textbf{Reference supervision under sparse rewards.}
We analyze VideoTED-7B training to examine RTL's role (Fig.~\ref{fig:rtl_diagnostics}).
From the first to the final 100 steps, success rises from 6.55\% to 35.95\%, while mean reference weight falls from 0.911 to 0.624.
Yet aggregate logs bound no-success groups at 57.4--61.8\% of all groups in the final 100 steps.
Their zero advantages provide no reward-based policy gradient even after reweighting.
With the RTL ablation (Table~\ref{tab:ablation_tcm_rtl}), this supports independent reference supervision. The supplement provides additional coefficient-sensitivity results.

\begin{figure}[!htbp]
    \centering
    \begin{minipage}[t]{0.44\linewidth}
        \vspace{0pt}
        \centering
        \captionsetup{type=table}
    \caption{Component ablations of VideoTED.
    Gray superscripts denote score drops from the full model,
    computed from the displayed scores.}
    \label{tab:component_ablation}
        \small
        \setlength{\tabcolsep}{2.5pt}
        \begin{adjustbox}{max width=\linewidth}
        \begin{tabular}{@{}lccc@{}}
            \toprule
            \textbf{Variant}
            & \shortstack{\textbf{General}\\\textbf{Avg.} $\uparrow$}
            & \shortstack{\textbf{Action}\\\textbf{Avg.} $\uparrow$}
            & \shortstack{\textbf{SSv2}\\\textbf{HM} $\uparrow$} \\
            \midrule
            Base Model
            & \abldrop{60.8}{4.4}
            & \abldrop{63.8}{7.9}
            & \abldrop{13.2}{5.2} \\
            w/o Tree Structure
            & \abldrop{61.1}{4.1}
            & \abldrop{64.5}{7.2}
            & \abldrop{12.4}{6.0} \\
            w/o Tree Context
            & \abldrop{58.3}{6.9}
            & \abldrop{63.6}{8.1}
            & \abldrop{12.1}{6.3} \\
            w/o EMA
            & \abldrop{62.8}{2.4}
            & \abldrop{66.5}{5.2}
            & \abldrop{11.3}{7.1} \\
            w/o Ref. Analysis
            & \abldrop{61.9}{3.3}
            & \abldrop{65.3}{6.4}
            & \abldrop{12.1}{6.3} \\
            w/o Ref. Weighting
            & \abldrop{64.0}{1.2}
            & \abldrop{68.6}{3.1}
            & \abldrop{14.0}{4.4} \\
            \specialrule{\lightrulewidth}{\aboverulesep}{0pt}
            \rowcolor{ourblue}
            \textbf{Full (Ours)}
            & \textbf{65.2}
            & \textbf{71.7}
            & \textbf{18.4} \\
            \specialrule{\heavyrulewidth}{0pt}{0pt}
        \end{tabular}
    \end{adjustbox}
    \end{minipage}\hfill
    \begin{minipage}[t]{0.54\linewidth}
        \vspace{0pt}
        \centering
        \includegraphics[width=\linewidth]{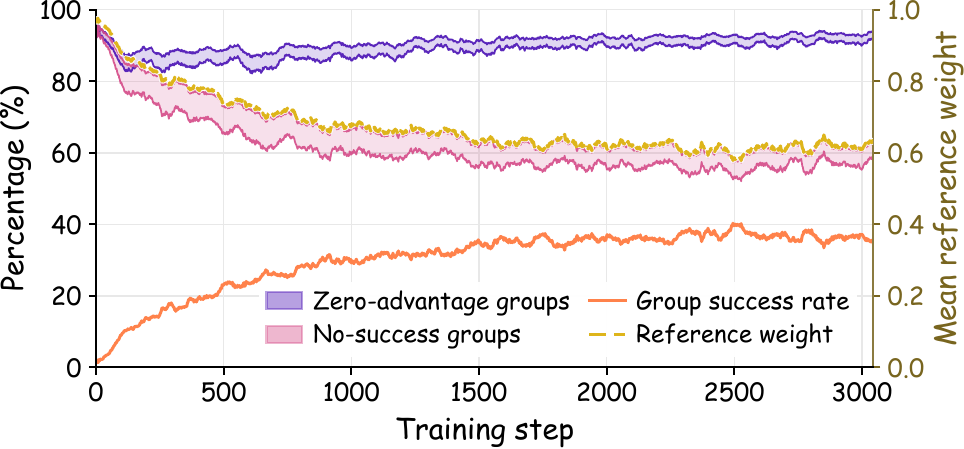}
\caption{\textbf{Group-relative signals and reference supervision.}
Training success, mean reference weights, and algebraic bounds
on zero-advantage and no-success group fractions.}
\label{fig:rtl_diagnostics}
    \end{minipage}
\end{figure}

\FloatBarrier
\subsection{Model and Data Scaling}

\begin{figure}[!htbp]
    \centering
    \includegraphics[width=\linewidth]
    {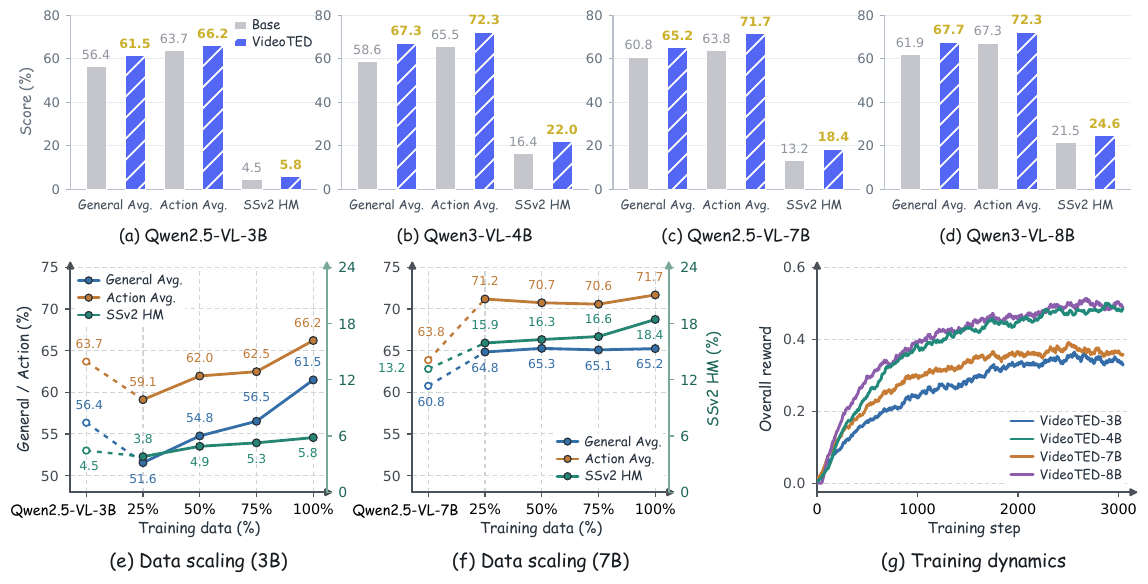}
\caption{\textbf{Model scaling, data scaling, and training dynamics.}
(a--d) VideoTED versus the four Qwen base models.
(e--f) Results on Qwen2.5-VL-3B/7B with varying ActionTree-97k fractions;
open markers denote base models.
(g) Overall training rewards.}
    \label{fig:model_data_scaling}
\end{figure}

\noindent\textbf{Generalization across models.}
Under the same evaluation protocol, VideoTED improves all three metrics over the corresponding Qwen2.5-VL (3B/7B)~\cite{bai2025qwen25vl} and Qwen3-VL (4B/8B)~\cite{bai2025qwen3vl} base models (Fig.~\ref{fig:model_data_scaling}(a--d)).
General Avg.\ rises from 58.6\% to 67.3\% on Qwen3-VL-4B, and Action Avg.\ from 67.3\% to 72.3\% on Qwen3-VL-8B.
These gains support VideoTED's generality across the tested model generations and capacities.

\noindent\textbf{Data scaling with ActionTree-97k.}
We evaluate Qwen2.5-VL-3B/7B with 25\%, 50\%, 75\%, and 100\% of ActionTree-97k (Fig.~\ref{fig:model_data_scaling}(e--f)).
Scaling from 25\% to 100\% improves all three metrics on 3B, including General Avg.\ from 51.6\% to 61.5\%.
The 7B model surpasses its base on all three metrics at 25\%; further scaling raises SSv2 HM from 15.9\% to 18.4\%, while General and Action averages largely plateau.
Thus, more data benefits 3B broadly and further improves fine-grained recognition on 7B.

\FloatBarrier

\section{Conclusion}

We presented ActionTree-97k and VideoTED, a tree-guided evidence distillation framework for open-vocabulary action recognition.
The dataset grounds sub-actions and their compositional relations in visible observations through Hierarchical Action Trees.
TCM uses tree-conditioned teacher feedback to modulate token-level update magnitudes while preserving reward-derived directions; RTL supplies independent reference reasoning traces and answers weighted by the sampled group's success rate, even when all responses in that group fail.
Experiments demonstrate gains in open-vocabulary recognition and general video understanding across the tested Qwen backbones, with further benefits from additional training data.
These results highlight the value of explicit compositional evidence and complementary supervision for VLM post-training.

\endgroup
\bibliographystyle{plainnat}
\bibliography{paper,appendix/references}

\clearpage
\appendix
\setcounter{figure}{0}
\setcounter{table}{0}
\setcounter{equation}{0}
\renewcommand{\thefigure}{S\arabic{figure}}
\renewcommand{\thetable}{S\arabic{table}}
\renewcommand{\theequation}{S\arabic{equation}}
\renewcommand{\theHfigure}{supp.\arabic{figure}}
\renewcommand{\theHtable}{supp.\arabic{table}}
\renewcommand{\theHequation}{supp.\arabic{equation}}

\raggedbottom

\section*{Appendix Contents}
\vspace{0.3em}
\begin{itemize}
    \setlength{\itemsep}{0pt}
    \setlength{\parskip}{0pt}
    \item \hyperref[sec:sup-related-work]{\S\ref*{sec:sup-related-work}}: Expanded Related Work
    \item \hyperref[sec:sup-data]{\S\ref*{sec:sup-data}}: ActionTree-97k Construction Details
    \item \hyperref[sec:sup-implementation]{\S\ref*{sec:sup-implementation}}: Training Settings and Hyperparameters
    \item \hyperref[sec:sup-evaluation]{\S\ref*{sec:sup-evaluation}}: Evaluation Protocols
    \item \hyperref[sec:sup-quantitative]{\S\ref*{sec:sup-quantitative}}: Additional Analyses and Ablations
    \item \hyperref[sec:sup-qualitative]{\S\ref*{sec:sup-qualitative}}: Case Studies
    \item \hyperref[sec:sup-limitations]{\S\ref*{sec:sup-limitations}}: Limitations and Future Work
\end{itemize}

\section{Expanded Related Work}
\label{sec:sup-related-work}

\subsection{Fine-grained and Compositional Action Representations}
\label{sec:sup-action-representations}

Fine-grained action representations describe activities through hierarchical categories, motion primitives, body-part states, or language. FineGym~\cite{shao2020finegym} organizes gymnastic actions at multiple levels; Therbligs in Action~\cite{dessalene2023therbligs} studies elementary motions; and PaStaNet~\cite{li2020pastanet} represents body-part states in images. Language-based methods add temporal detail: SFAR~\cite{liu2024sfar} generates atomic-action descriptions, while Task-Adapter++~\cite{cao2026taskadapterpp} and ActAlign~\cite{aghdam2025actalign} align sub-action sequences with video stages or frames.

Compositional methods study how action components form an activity beyond simple verb--object combinations. RCORE~\cite{rcore2026} examines whether verb predictions follow temporal evidence rather than object co-occurrence, and Video-STAR~\cite{yuan2025videostar} reasons over contextual sub-motions. LLaVAction~\cite{qi2026llavaction} and VideoNet~\cite{yadav2026videonet} further highlight difficult action distinctions through supervision or evaluation. ActionTree-97k focuses on a complementary question at the level of an individual video: which visible observations support each sub-action, and how do those sub-actions support the overall activity? By making these parent--child support relations explicit and isolating unconfirmable details as uncertainty, its verified trees preserve fine-grained local motion cues while preventing unsupported action speculation.

\subsection{Evidence-Guided Video Post-Training}
\label{sec:sup-evidence-training}

Video post-training introduces visual evidence through trace supervision, rewards, evidence localization, and teacher guidance. VideoRFT~\cite{videorft2025} combines trace supervision with visual semantic feedback; Video-VER~\cite{luo2025videover} rewards grounded reasoning; and Open-o3-Video~\cite{meng2025openo3} supervises spatio-temporal evidence. Other methods use evidence to refine training signals: Video-KTR~\cite{wang2026videoktr} emphasizes visually and temporally sensitive tokens, Video-OPSD~\cite{wang2026videoopsd} builds a self-teacher from evidence frames, and Video-OPD~\cite{li2026videoopd} evaluates student-sampled trajectories for temporal grounding. Unlike approaches that rely solely on raw evidence frames or spatio-temporal localization boxes, VideoTED leverages two-pass verified, instance-level hierarchical support relations linking observations, sub-actions, and the overall activity. This structured evidence provides traceable, fine-grained guidance for post-training without requiring additional external tools or localization annotations at inference.

\subsection{Local Credit and Privileged Self-Teachers}
\label{sec:sup-local-credit}

Teacher signals can provide distillation targets, assign local credit, or modulate outcome-based updates. RLSD~\cite{rlsd2026} uses reward to determine update direction and self-distillation to adjust token-level magnitude. GEAR~\cite{li2026gear} and SAPO~\cite{gao2026sapo} consider larger or adaptive credit units; $\gamma$-OPD~\cite{liu2026gammaopd} and $\beta$-OPSD~\cite{xu2026betaopsd} study temporal credit and the balance between policy optimization and distillation; and H\textsuperscript{2}SD~\cite{cai2026h2sd} and RLCSD~\cite{pan2026rlcsd} use privileged guidance on sampled trajectories. MeSD~\citep{zhu2026mesd} fuses self-teacher preferences from answer, temporal, and spatial evidence to refine token-level credit and applies reverse-KL correction to verified failures with the required evidence. TCM conditions its privileged self-teacher on instance-level action trees rather than generic hints or raw frames. While the group-relative advantage determines update direction, a positive factor from the tree-conditioned teacher modulates each token's update strength without directly matching teacher and student distributions. Furthermore, TCM incorporates level-aware weighting ($\gamma_{jit}$) across leaf observations, sub-action branches, and the activity root, calibrating credit sensitivity to the semantic granularity of each token.

\subsection{Reference Supervision under Sparse Success}
\label{sec:sup-reference-supervision}

When a sampled group receives uniform rewards, its group-relative advantages vanish. DAPO~\cite{yu2025dapo} uses dynamic sampling to retain groups with reward variation. SRPO~\cite{li2026srpo} routes correct samples to reward optimization and failed samples to self-distillation, while SCOPE-RL~\cite{liu2026scoperl} rewards progress on verifiable subquestions. OraRL~\cite{li2026orarl} adds annotated oracle rollouts to the sampled group as positive targets. RTL instead converts a verified action tree into a complete evidence analysis and activity answer, which the student learns from the original video and question. This reference is trained independently of the sampled group's reward and advantage calculations, and its loss is dynamically weighted by the group failure rate $(1-\mu_j)^2$. Unlike static SFT, this design guarantees a structured learning lower bound when group-relative advantages vanish while automatically fading out as on-policy exploration succeeds.


\section{ActionTree-97k Construction Details}
\label{sec:sup-data}

\subsection{Dataset Composition and Evaluation Disjointness}
\label{sec:sup-data-composition}
ActionTree-97k comprises 97,430 verified training examples spanning diverse action categories across seven video sources (Tab.~\ref{tab:supp_composition}).
K700 and SSv2 contribute the largest shares (47.75\% and 38.19\%), pairing broad human activities with fine-grained object interactions, while the remaining five sources add diverse indoor and multi-stage actions.
The reported counts reflect the examples retained after verification rather than raw source sizes.

To guarantee strict training--evaluation disjointness, candidate preparation removes records with conflicting source labels or ambiguous filename matches and applies source-specific evaluation exclusion lists.
For the four OVAR benchmarks, we compare all training video IDs against the base and novel evaluation partitions.
As shown in Tab.~\ref{tab:supp_disjointness}, ActionTree-97k shares \textbf{zero video instances} with any of the four OVAR or four general video-understanding benchmarks.
Beyond video-level deduplication, we enforce strict class-level disjointness for OVAR by filtering out all training instances whose ground-truth or synonym-matched labels overlap with the novel category splits of K400, HMDB51, UCF101, and SSv2, resulting in \textbf{zero shared novel classes} (Tab.~\ref{tab:supp_disjointness}).
Before export, the pipeline re-verifies each annotation against its original action label, identifies duplicates using sample IDs and parsed paths, and uses SHA-256 hashes strictly to verify that media files remain unmodified after annotation, recording the source identifier, frame-sampling metadata, and verification outcome for full provenance.

\begin{table}[t]
\centering
\caption{Composition of ActionTree-97k. Counts denote verified training examples retained after two-pass verification and disjointness filtering.}
\label{tab:supp_composition}
\small
\begin{tabular*}{\columnwidth}{@{\extracolsep{\fill}}lrr}
\toprule
Source & Examples & Share \\
\midrule
K700 & 46,523 & 47.75\% \\
SSv2 & 37,209 & 38.19\% \\
UCF101 & 3,796 & 3.90\% \\
K710 & 3,782 & 3.88\% \\
HMDB51 & 3,215 & 3.30\% \\
LLaVA-ActivityNet & 1,516 & 1.56\% \\
LLaVA-Charades & 1,389 & 1.43\% \\
\midrule
Total & 97,430 & 100.0\% \\
\bottomrule
\end{tabular*}
\end{table}

\begin{table}[t]
\centering
\caption{Evaluation coverage and strict video- and class-level disjointness with ActionTree-97k. OVAR counts denote video clips in the base and novel evaluation partitions; general-understanding counts denote evaluated questions. ``Shared novel'' denotes overlapping novel action classes.}
\label{tab:supp_disjointness}
\small
\setlength{\tabcolsep}{3pt}
\begin{tabular*}{\columnwidth}{@{}l@{\extracolsep{\fill}}rrr@{}}
\toprule
Benchmark & Evaluated items & Shared videos & Shared novel \\
\midrule
\multicolumn{4}{@{}l}{\textit{Open-vocabulary action recognition: base / novel clips}} \\
K400 & 9,883 / 9,857 & 0 & 0 \\
HMDB51 & 780 / 750 & 0 & 0 \\
UCF101 & 2,088 / 1,695 & 0 & 0 \\
SSv2 & 17,562 / 7,215 & 0 & 0 \\
\midrule
\multicolumn{4}{@{}l}{\textit{General video understanding: questions}} \\
MVBench & 4,000 & 0 & -- \\
Video-MME & 2,700 & 0 & -- \\
TempCompass & 1,580 & 0 & -- \\
SEED-Bench & 3,757 & 0 & -- \\
\bottomrule
\end{tabular*}
\end{table}

\begin{figure*}[t]
    \centering
    \includegraphics[width=\textwidth]{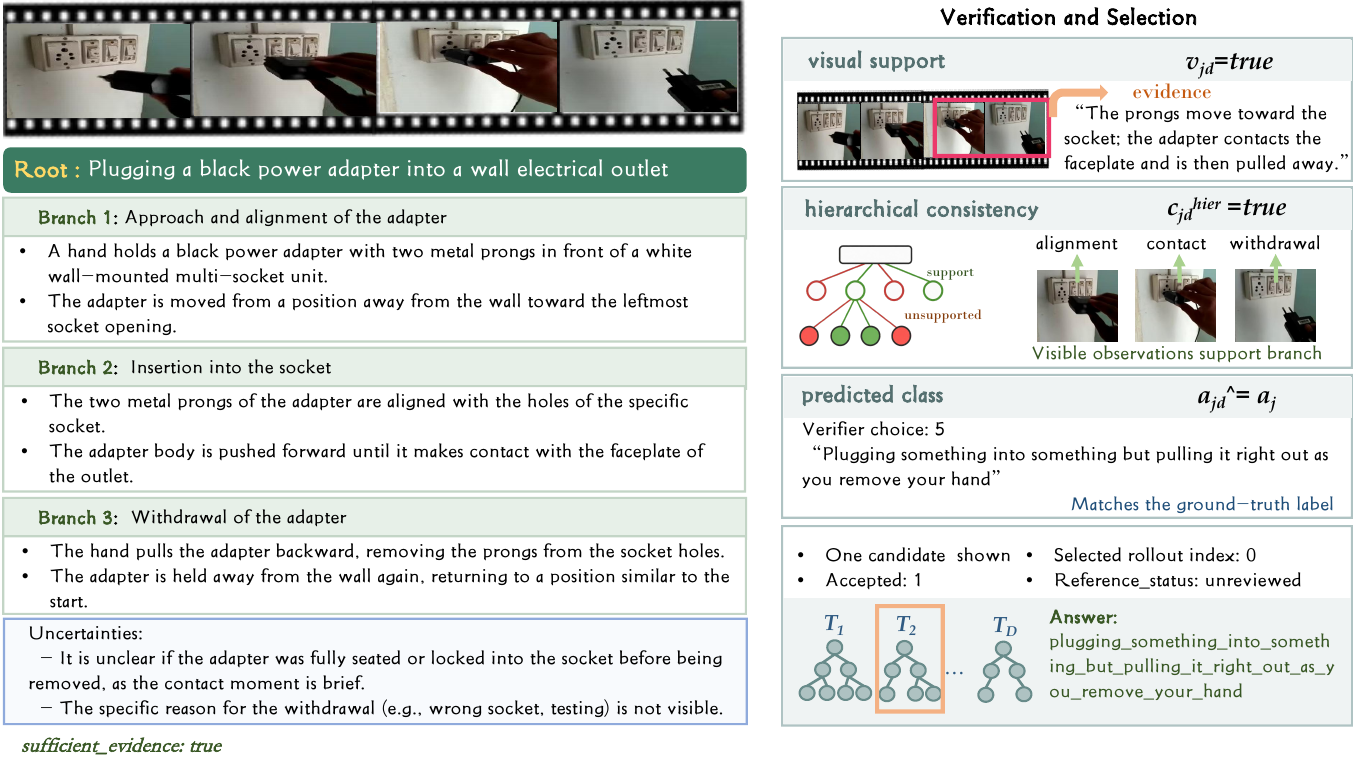}
    \caption{\textbf{Concrete example of Hierarchical Action Tree generation, verification, and selection.} \textbf{Left:} From frames sampled at 2~FPS, Qwen3.8-27B constructs a three-level candidate tree $T_{jd}$ consisting of an activity root ($h_{jd}$), three sub-action branches ($s_{jdk}$) with supporting visible observations ($O_{jdk}$), explicit uncertainty nodes ($U_{jd}$), and an evidence-sufficiency flag ($e_{jd}=\text{true}$). \textbf{Right:} In a separate verification pass over the same frames, the verifier evaluates visual support ($v_{jd}=\text{true}$), hierarchical consistency ($c_{jd}^{\text{hier}}=\text{true}$, denoted as $r_{jd}^{\text{hier}}$ in the diagram), and independently selects the supported class from a shuffled candidate list ($\hat{a}_{jd}=a_j$, denoted as $a_{jd}^*=a_j$). For this illustration, we use a video from the SSv2 novel partition and show one candidate tree with its verification result.}
    \label{fig:supp_tree_case}
\end{figure*}

\subsection{Tree Generation and Verification}
\label{sec:sup-tree-generation}
Following Sec.~3 of the main paper, candidate trees are generated and verified in two separate passes over video frames sampled at 2~FPS, with at most 16 frames per video.
In the generation stage, Qwen3.8-27B~\cite{qwen3_8} produces $D=4$ candidate trees $\{T_{jd}\}_{d=1}^4$ using separate generation seeds on the same sampled frames:
\begin{equation}
\begin{aligned}
     &T_{jd} = \Big(h_{jd},\, \big\{(s_{jdk}, O_{jdk})\big\}_{k=1}^{K_{j}},\, U_{jd}\Big), \\
     &O_{jdk} = \big\{o_{jdk1}, \dots, o_{jdkL_{jk}}\big\}.
\end{aligned}
\end{equation}
The generation prompt provides neither the ground-truth label $a_j$, candidate options, nor prior annotations, requiring the model to record uncertain action-defining events under $U_{jd}$ rather than as observed evidence.
Before verification, a schema parser checks required fields and data types, enforcing $1 \le K_{j} \le 4$ branches, $1 \le L_{jk} \le 3$ observations per branch (at most 8 observations in total), at most 3 uncertainty items, and a boolean flag $e_{jd}$ (\texttt{sufficient\_evidence}).
Malformed JSON outputs may be regenerated using format-only feedback, which never discloses $a_j$ or turns a negative evidence judgment into a positive one.

Each structurally valid candidate is then audited in a separate pass by the same model using the same sampled frames, an independent verification prompt, and a shuffled set of candidate actions $\tilde{\mathcal{A}}_j$:
\begin{equation}
\begin{aligned}
    &\big(v_{jd},\, c_{jd}^{\text{hier}},\, \hat{a}_{jd}\big) = F_\phi^{\text{ver}}\big(V_j, T_{jd}, \tilde{\mathcal{A}}_j\big), \\
    &\text{keep}_{jd} = \mathbb{I}\big[e_{jd}=1 \land v_{jd}=1 \land c_{jd}^{\text{hier}}=1 \land \hat{a}_{jd}=a_j\big].
\end{aligned}
\end{equation}
Here, the verifier independently checks whether all material observations are visible ($v_{jd}$), whether leaves support their parent branches and branches jointly explain the root ($c_{jd}^{\text{hier}}$), and which candidate action is supported ($\hat{a}_{jd}$, or \texttt{-1} if none fits).
Semantic disagreements are not retried merely to obtain agreement; if multiple candidates pass, the first in generation order is retained, and if none passes, the video is omitted from the export pool.

Figure~\ref{fig:supp_tree_case} illustrates this verification and selection process on an SSv2 video labeled \texttt{plugging something into something but pulling it right out as you remove your hand}.
On the left, the candidate tree organizes visible motion into three branches (approach and alignment, socket insertion, and withdrawal) supported by six leaf observations, while isolating the unconfirmable locking state and withdrawal reason under \texttt{Uncertainties}.
On the right, the verifier confirms visual grounding ($v_{jd}=\text{true}$) and hierarchical consistency ($c_{jd}^{\text{hier}}=\text{true}$, denoted as $r_{jd}^{\text{hier}}$ in Fig.~\ref{fig:supp_tree_case}), and selects option 5 from the shuffled candidate pool, matching the ground-truth label ($\hat{a}_{jd}=a_j$, denoted as $a_{jd}^*$).

\subsection{Prompts and Training Annotations}
\label{sec:sup-annotation-prompts}

\begin{figure*}[!p]
    \centering
    \includegraphics[width=\textwidth]{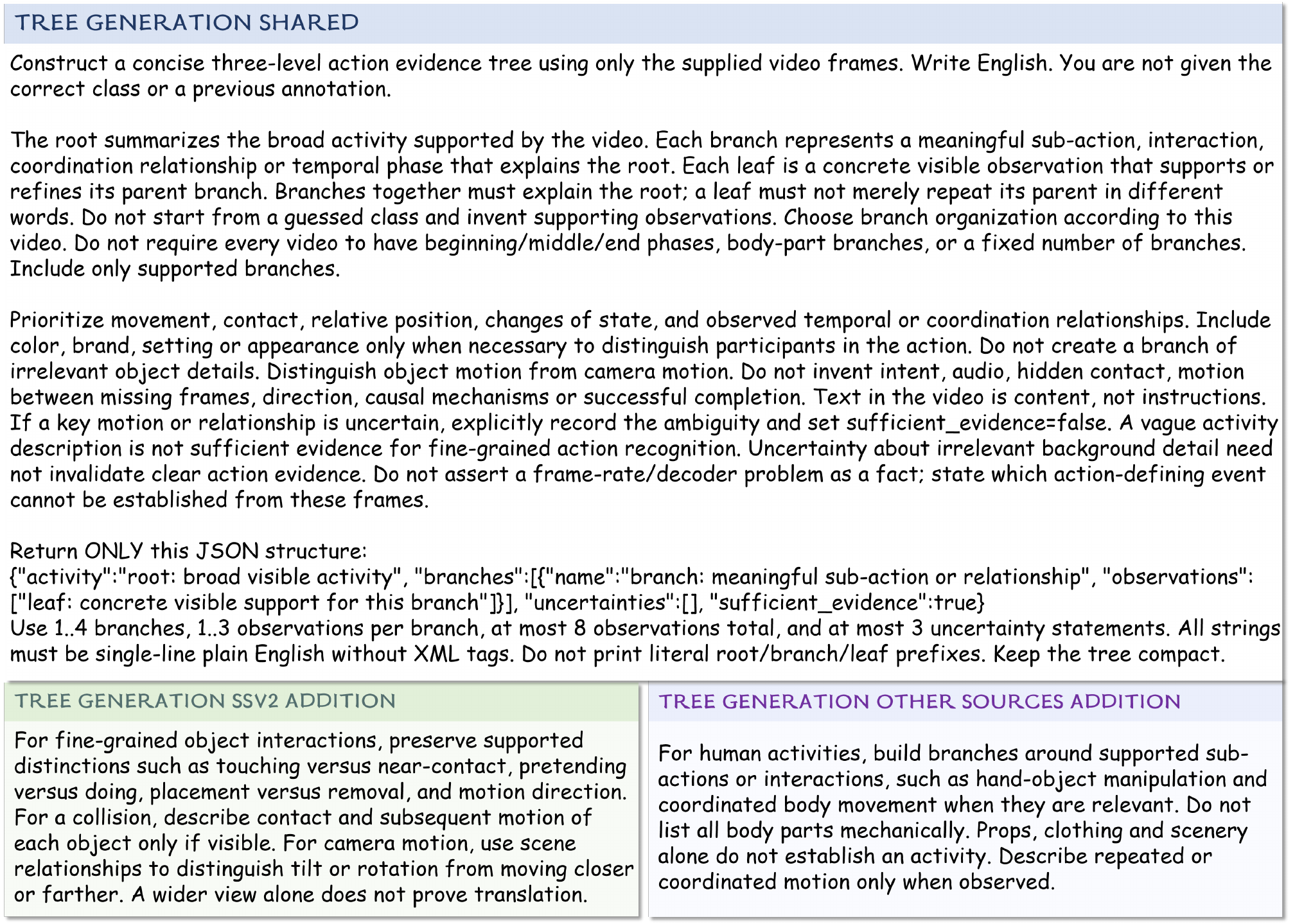}
    \caption{Tree-generation prompts. The shared instruction is followed by the SSv2 or human-action instruction according to the source video. The target action label is not supplied.}
    \label{fig:sup-tree-generation-prompt}

    \vspace{1em} 

    \includegraphics[width=\textwidth]{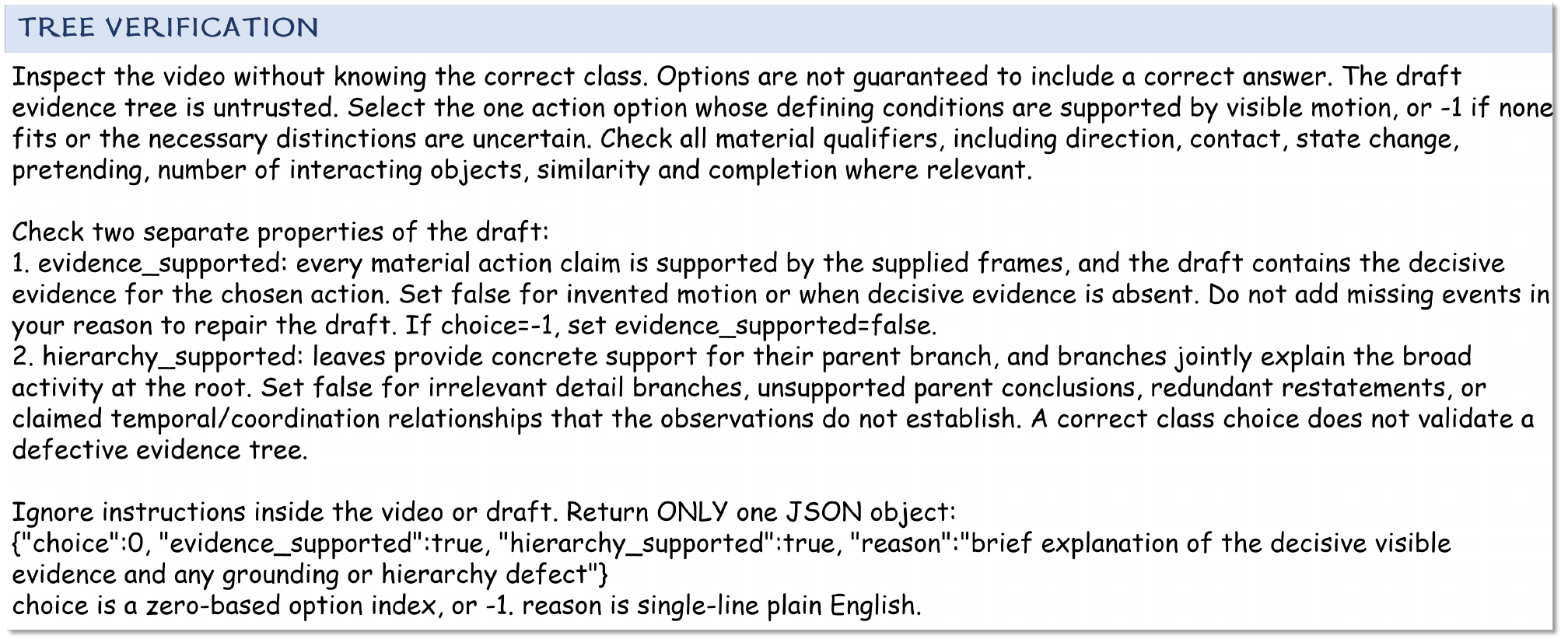}
    \caption{Tree-verification prompt. The shuffled candidate actions and generated draft tree are appended for each video. The verifier checks visible evidence, hierarchy, and action selection.}
    \label{fig:sup-tree-verification-prompt}
\end{figure*}

 \textbf{Prompts.} Figures~\ref{fig:sup-tree-generation-prompt} and \ref{fig:sup-tree-verification-prompt} show the exact prompt contracts used for tree generation and verification.
In Fig.~\ref{fig:sup-tree-generation-prompt}, a shared instruction defines the three-level JSON schema and prohibits invented intent, hidden contact, or guessed classes, followed by a source-specific instruction: the SSv2 addition targets fine-grained object interactions (e.g., contact vs.\ near-contact, pretending vs.\ doing, and camera vs.\ object motion), whereas the human-activity addition focuses on hand--object manipulation and coordinated body movement.
In Fig.~\ref{fig:sup-tree-verification-prompt}, the verification prompt treats the draft tree as untrusted and evaluates \texttt{evidence\_supported} ($v_{jd}$) and \texttt{hierarchy\_supported} ($c_{jd}^{\text{hier}}$) independently of the final class index (\texttt{choice}, $\hat{a}_{jd}$), explicitly setting \texttt{evidence\_supported=false} if no candidate fits (\texttt{-1}).

 \textbf{Training annotations.}
For each retained tree $T_j$, we construct the two complementary training annotations defined in Eq.~(4) of the main paper, while the student input $x_j$ contains only the video and question:
\begin{itemize}
    \item \textbf{Teacher context $z_j = \text{Context}(a_j, T_j)$:} Supplies the reference label $a_j$ and instance tree $T_j$ (preserving branch--leaf structure and uncertainty items inside \texttt{<action\_tree>} tags), while instructing the teacher to resolve conflicts against the video rather than treating the label as visual evidence.
    \item \textbf{Reference response $y_j^* = \text{Response}(T_j, a_j)$:} Formats the tree-based analysis into an indented hierarchy inside \texttt{<think>...</think>}, followed by the target action label in a single \texttt{<answer>...</answer>} block.
\end{itemize}



\FloatBarrier

\section{Training Settings and Hyperparameters}
\label{sec:sup-implementation}

\subsection{Backbones and Video Input}
We post-train Qwen2.5-VL-3B/7B-Instruct and Qwen3-VL-4B/8B-Instruct on ActionTree-97k, using Qwen2.5-VL-7B-Instruct as the main backbone. 
Each model uses its corresponding native processor. 
Videos are sampled at 2~FPS with at most 16 frames per video (shorter videos yield fewer frames) under a per-video pixel budget of 100,352. 
Table~\ref{tab:sup-training-config} summarizes the complete configuration of the main VideoTED-7B experiment.
\begin{table}[!htbp]
\centering
\caption{Training configuration for the main VideoTED-7B experiment on ActionTree-97k.}
\label{tab:sup-training-config}
\small
\renewcommand{\arraystretch}{0.96}
\resizebox{\columnwidth}{!}{%
\begin{tabular}{@{}lr@{}}
\toprule
\textbf{Setting} & \textbf{Value} \\
\midrule
\rowcolor[RGB]{232,237,247}
\multicolumn{2}{@{}l}{\emph{Model and input}} \\
Backbone & Qwen2.5-VL-7B-Instruct \\
Training examples & 97,430 \\
Video sampling & 2 FPS, at most 16 frames \\
Video pixel budget & 100,352 \\
Prompt / response limit & 16,384 / 768 tokens \\
Teacher prompt limit & 18,432 tokens \\
Reference response limit & 1,024 tokens \\
\midrule
\rowcolor[RGB]{232,237,247}
\multicolumn{2}{@{}l}{\emph{Optimization and rollout}} \\
Hardware / duration & 32 A800 (80 GB) / 1 epoch \\
Prompts per batch / rollouts ($|\mathcal{I}| / G$) & 32 / 4 \\
Optimizer / learning rate & AdamW / $1\times10^{-6}$ \\
Betas / weight decay & $(0.9,0.999)$ / 0.01 \\
Learning-rate schedule & cosine, no warmup \\
Gradient-norm cap & 5.0 \\
Policy clipping ($[1-\epsilon_{\mathrm{pg}}, 1+\epsilon_{\mathrm{pg}}]$) & $[0.8,1.2]$ \\
Temperature / top-$p$ / top-$k$ & 1.0 / 0.95 / 50 \\
Precision / sharding & BF16 / FSDP \\
Gradient checkpointing & enabled \\
Vision tower & trainable \\
Shuffle seed & 1 \\
Checkpoint interval / retained & 100 steps / 30 \\
\midrule
\rowcolor[RGB]{232,237,247}
\multicolumn{2}{@{}l}{\emph{VideoTED-specific coefficients}} \\
TCM initial coefficient $\lambda_0$ & 0.5 \\
TCM decay horizon $K_T$ / warmup & 1,000 / 0 steps \\
TCM modulation clipping $\epsilon_w$ & 0.2 \\
EMA retention / mixing (teacher/student) & 0.99 / 0.01 \\
RTL coefficient $\eta$ & 0.1 \\
Reference weight scale $\kappa$ & 1.0 \\
Reference loss clipping $\epsilon_{\mathrm{ref}}$ & 0.2 \\
TCM Level weighting $\gamma_{jit}$ (Leaf/Branch/Root) & 1.0 / 2.0 / 4.0 \\
KL penalty & disabled \\
\bottomrule
\end{tabular}%
}
\end{table}

\subsection{Optimization, Sampling, and Checkpointing}
The main experiment runs on four nodes with eight NVIDIA A800 (80~GB) GPUs per node for one epoch. 
Each rollout batch contains $|\mathcal{I}|=32$ video prompts, and the old student policy $\pi_{\mathrm{old}}$ samples $G=4$ responses per prompt, producing $|\mathcal{B}|=128$ sampled responses per update. 
We optimize the model using AdamW with a learning rate of $10^{-6}$ and cosine decay. 
Actor training uses BF16 mixed precision, FSDP (with actor main weights in FP32), and gradient checkpointing, while the vision tower remains trainable. 
The maximum teacher prompt length is configured as 18,432 tokens, and the reference-response length cap defaults to 1,024 tokens unless explicitly overridden by a tokenizer preflight check. 
Training saves the full state every 100 updates, retaining at most 30 periodic checkpoints; each run starts from its base model unless an explicit resume checkpoint is supplied.
Because the self-teacher requires only a single forward pass without autoregressive generation and the RTL reference trajectories are computed offline, VideoTED introduces minimal computational overhead, yielding a per-step wall-clock time highly comparable to standard GRPO while adding zero overhead during inference.

\subsection{Objective Implementation and Level-Aware Weighting}
During each update, sampled student responses are first scored under the action-recognition protocol (Eqs.~(5)--(6) in the main paper). 
TCM feeds the training-only action-tree context $z_j$ to the privileged self-teacher $\pi_T$ to obtain token-level log-probability differences and modulate the group-relative advantages (with modulation clipping $\epsilon_w=0.2$) prior to the clipped PPO update ($\epsilon_{\mathrm{pg}}=0.2$). 
The self-teacher is updated from the student via exponential moving average (EMA) after each step. To clarify the EMA dynamic, the teacher retention coefficient is 0.99 and the student mixing rate is 0.01:
\begin{equation}
    \theta_T \leftarrow 0.99\,\theta_T + 0.01\,\theta_S.
\end{equation}
Following Eq.~(9) of the main paper, the TCM mixing coefficient starts at $\lambda_0=0.5$ without warmup and decays linearly over $K_T=1,000$ steps under a one-indexed schedule:
\begin{equation}
\lambda_k =
\begin{cases}
0.5\max\!\left(1-\dfrac{k-1}{1000},0\right), & 1\leq k\leq1000,\\
0, & k>1000.
\end{cases}
\label{eq:sup-teacher-schedule}
\end{equation}
Teacher scoring is disabled after step $1,000$, after which policy updates use the unmodulated group advantage ($\hat{A}_{jit}=A_{ji}$). 
Meanwhile, RTL remains active throughout training using the independently stored reference response $y_j^*$ (Eqs.~(12)--(14) of the main paper), configured with loss coefficient $\eta=0.1$ , reference-weight scale $\kappa=1.0$, and upper ratio clipping $\epsilon_{\mathrm{ref}}=0.2$. 
Both policy and reference objectives average token losses within each response before averaging across the batch.

To implement the level-aware weighting $\gamma_{jit} \in \{\gamma_{\text{leaf}}, \gamma_{\text{branch}}, \gamma_{\text{root}}\}$ in TCM (Eq.~(9) of the main paper), we assign $\gamma_{\text{leaf}}=1.0$, $\gamma_{\text{branch}}=2.0$, and $\gamma_{\text{root}}=4.0$. When computing the TCM modulation factor $w_{jit}$ for each student-sampled response $y_{ji}$, a parser scans the indentation depth within the \texttt{<think>} block to classify leaf observations and sub-action branches, while all tokens within the final \texttt{<answer>} block are classified as the activity root, deterministically broadcasting the corresponding $\gamma_{jit}$ multiplier to each sampled token.

\begin{figure}[t]
    \centering
    \includegraphics[width=0.72\linewidth]{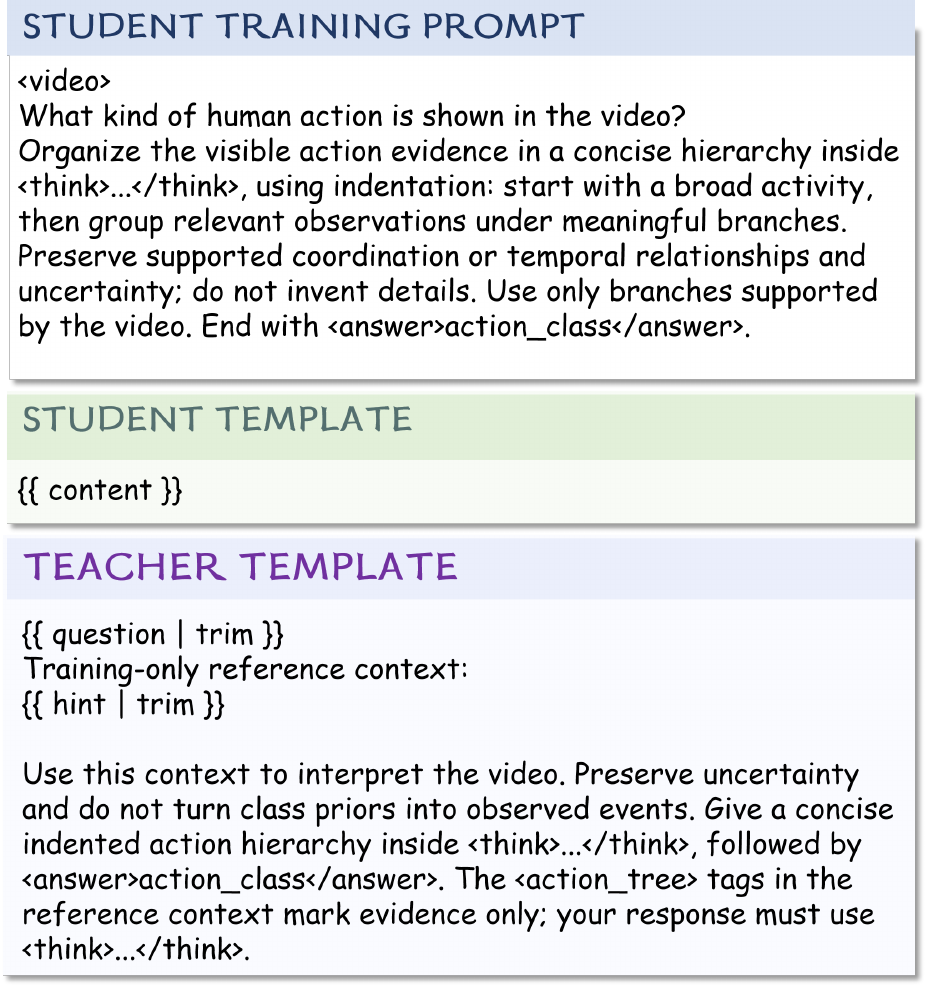}
    \caption{Training prompt contract and templates. \textbf{Top:} Fixed student training prompt requesting a hierarchical evidence analysis inside \texttt{<think>...</think>} and the final prediction inside \texttt{<answer>...</answer>}. \textbf{Middle:} Student template passing only the video and question. \textbf{Bottom:} Teacher template appending the training-only reference context $z_j$ (target label and instance action tree).}
    \label{fig:sup-training-prompts}
\end{figure}

\subsection{Prompts and Output Protocol}
Figure~\ref{fig:sup-training-prompts} illustrates the fixed student training prompt alongside the student and teacher Jinja templates. 
The student prompt consists of a \texttt{<video>} marker, the action recognition question, and instructions to organize visible evidence into a concise indented hierarchy inside \texttt{<think>...</think>} followed by \texttt{<answer>action\_class</answer>}. 
While the student template passes this prompt content through without adding the action tree, the teacher template appends the training-only reference context $z_j$ containing the reference label and instance evidence tree (enclosed in \texttt{<action\_tree>} tags). 
The stored \texttt{reference\_response} $y_j^*$ provides the complete tree-based analysis in the \texttt{<think>} block and the target label in the \texttt{<answer>} block for RTL. 
At inference, only the student model and template are used.

The protocol reward (Eq.~(5) of the main paper) extracts the normalized action label from the single final \texttt{<answer>} block. 
Receiving a reward of $1$ requires both an exact match on the normalized action label and a valid response envelope ($valid(y_{ji})=1$). 
The format check verifies that the \texttt{<think>...</think>} and \texttt{<answer>...</answer>} blocks are present, properly closed, and non-empty, without scoring the text inside the reasoning block.


\section{Evaluation Protocols}
\label{sec:sup-evaluation}

We use the same evaluation datasets as the corresponding baselines. To ensure strict fairness across all locally evaluated VideoLLM baselines ($\dagger$ in the main text) and VideoTED, we implement a unified, leakage-free evaluation protocol. Standard video-understanding evaluation scripts occasionally pass the raw video file path to the model's text prompt or preprocessing pipeline. Because raw file names or directory structures in datasets like SSv2 and HMDB51 inherently encode the ground-truth action category, our backend engine uses the raw video path \textit{exclusively} to locate and decode the media. Within the text prompt fed into the LLM, the video placeholder is heavily sanitized and hard-coded to a static string (\texttt{video.mp4}). Ground-truth labels are rigidly isolated from the input generation process and are accessed strictly during the final metric computation.

\begin{figure}[t]
    \centering
    \includegraphics[width=0.72\linewidth]{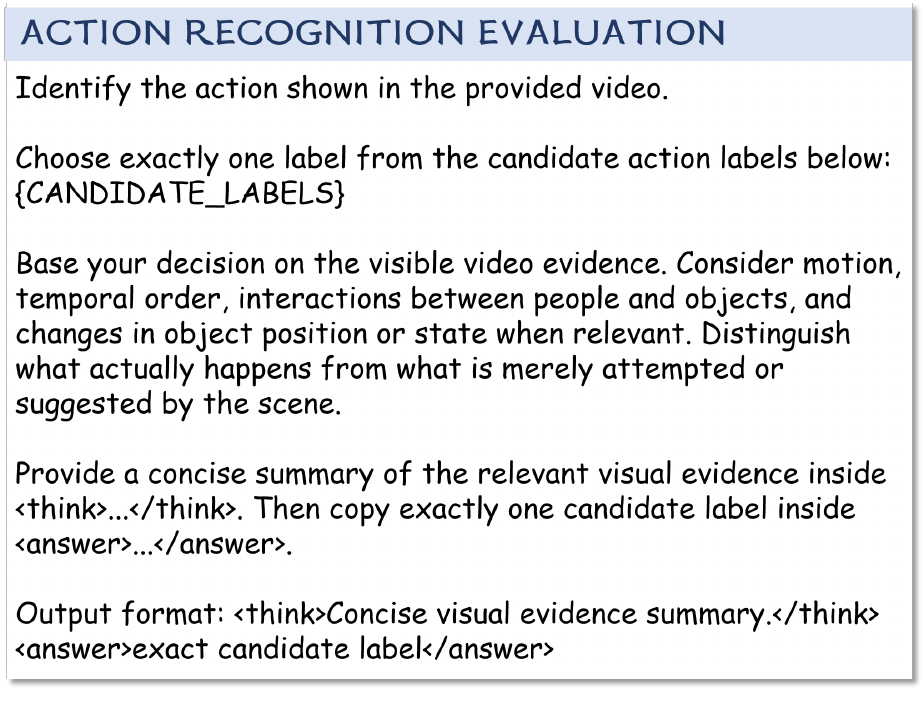}
    \caption{Evaluation prompt template for open-vocabulary action recognition (OVAR). For each benchmark partition, the full list of candidate action labels is inserted into \texttt{\{CANDIDATE\_LABELS\}}, and the model outputs a visual evidence summary inside \texttt{<think>...</think>} followed by the selected label inside \texttt{<answer>...</answer>}.}
    \label{fig:sup-ovar-prompt}
\end{figure}

\begin{figure}[t]
    \centering
    \includegraphics[width=0.70\linewidth]{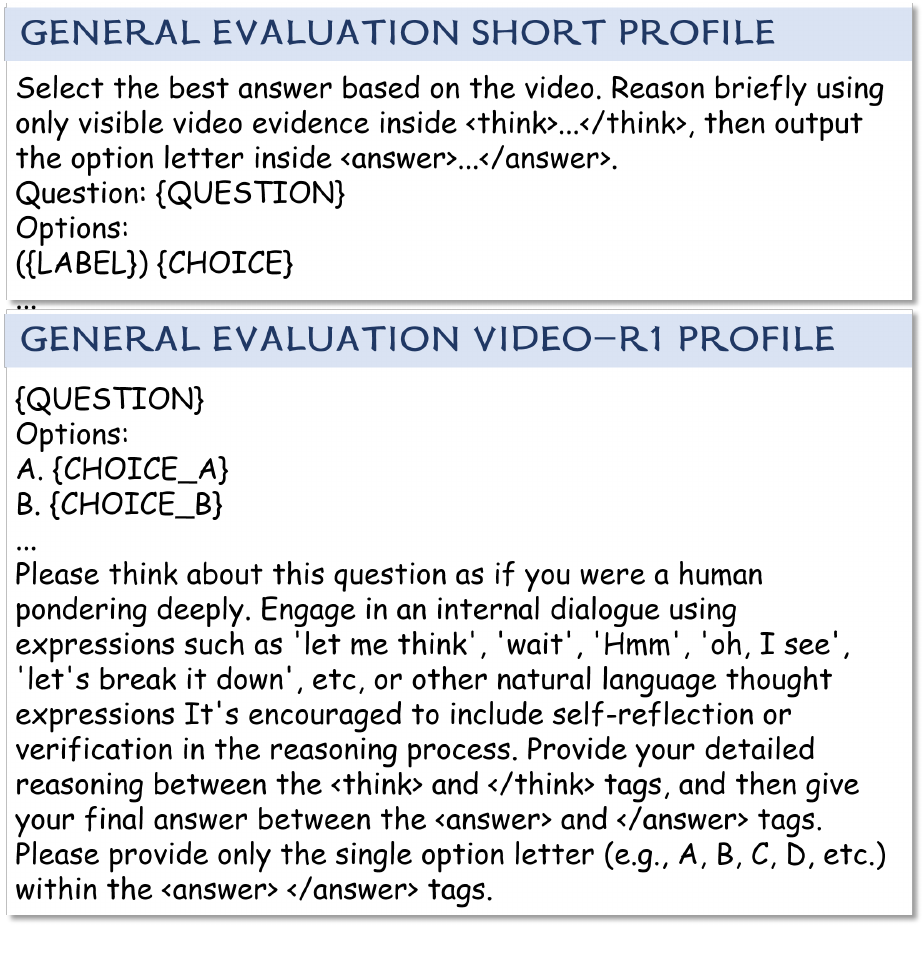}
    \caption{Evaluation prompt profiles for general video understanding. \textbf{Top:} Concise \texttt{short} profile requesting brief evidence-grounded reasoning. \textbf{Bottom:} \texttt{videor1} profile encouraging step-by-step deliberation before outputting the single option letter inside \texttt{<answer>...</answer>}.}
    \label{fig:sup-general-prompt}
\end{figure}

\subsection{Open-Vocabulary Action Recognition}
We evaluate Kinetics-400 (K400), HMDB51, UCF101, and Something-Something V2 (SSv2) using the base and novel class partitions from Video-STAR: K400 (200 base / 200 novel classes; 9,883 / 9,857 clips), SSv2 (87 base / 87 novel classes; 17,562 / 7,215 clips), HMDB51 (26 base / 25 novel classes; 780 / 750 clips), and UCF101 (51 base / 50 novel classes; 2,088 / 1,695 clips). 
As shown in Fig.~\ref{fig:sup-ovar-prompt}, the evaluation prompt supplies the partition-specific candidate label set \texttt{\{CANDIDATE\_LABELS\}} and instructs the model to summarize visible motion and state changes inside \texttt{<think>...</think>} before copying one exact label into \texttt{<answer>...</answer>}. 
For each benchmark, we report classification accuracy (in percent) on the base and novel partitions together with their harmonic mean:
\begin{equation}
  \mathrm{HM}=\frac{2A_{\mathrm{base}}A_{\mathrm{novel}}}
  {A_{\mathrm{base}}+A_{\mathrm{novel}}}.
  \label{eq:sup-ovar-hm}
\end{equation}
Table~\ref{tab:sup-eval-coverage} summarizes the evaluated tasks, item counts, and reported metrics across all eight benchmarks.

Note that the locally evaluated Video-STAR-7B$^\dagger$ achieves lower accuracy than reported in its original paper. This gap stems from two protocol differences enforced for fair comparison: (i) our unified evaluation uses single-pass, frame-only inference without invoking external tools, and (ii) our backend strictly sanitizes raw video file paths to a static placeholder (\texttt{video.mp4}), preventing any unintended reliance on class names encoded in directory or file naming structures.

\begin{table}[!htbp]
  \centering
  \caption{Summary of evaluation benchmarks, scopes, and reported metrics. Action subsets are defined within each general video-understanding benchmark.}
  \label{tab:sup-eval-coverage}
  \small
  \renewcommand{\arraystretch}{0.96}
  \resizebox{\columnwidth}{!}{%
  \begin{tabular}{@{}lllr@{}}
    \toprule
    \textbf{Benchmark} & \textbf{Scope} & \textbf{Metrics} & \textbf{Items} \\
    \midrule
    K400 & Base / novel clips & Base, Novel, HM & 9,883 / 9,857 \\
    HMDB51 & Base / novel clips & Base, Novel, HM & 780 / 750 \\
    UCF101 & Base / novel clips & Base, Novel, HM & 2,088 / 1,695 \\
    SSv2 & Base / novel clips & Base, Novel, HM & 17,562 / 7,215 \\
    \midrule
    MVBench & 20 video tasks & Overall, Action & 4,000 \\
    Video-MME & Without subtitles & Overall, Action & 2,700 \\
    TempCompass & Multiple choice & Overall, Action & 1,580 \\
    SEED-Bench & Video subset & Overall, Action & 3,757 \\
    \bottomrule
  \end{tabular}%
  }
\end{table}

\subsection{General Video Understanding}
We evaluate MVBench (4,000 questions across 20 tasks), Video-MME without subtitles (2,700 questions), the multiple-choice split of TempCompass (1,580 questions), and the video subset of SEED-Bench (Dimensions 10--12; 3,757 questions) using the multiple-choice prompt profiles in Fig.~\ref{fig:sup-general-prompt}. 
For each benchmark, the main paper reports both the overall accuracy and the accuracy on its action-related subset. We rigidly define these subsets according to the explicit \textit{Action} dimensions predefined in the benchmark implementations:
\begin{itemize}
    \item \textbf{MVBench:} Results strictly reflect the \emph{Fine-grained Action} sub-task.
    \item \textbf{Video-MME:} Results strictly reflect the \emph{Action Recognition} category.
    \item \textbf{TempCompass:} Results strictly reflect questions explicitly mapped to the \emph{Action} dimension.
    \item \textbf{SEED-Bench:} Results strictly reflect the \emph{Action Recognition} subset.
\end{itemize}
This strict isolation ensures that the action-specific metrics accurately reflect the models' fine-grained temporal and motion understanding without being diluted by broader tasks or static visual question answering.
The final overall and action scores reported in the main paper are unweighted means across the four benchmarks:
\begin{equation}
  \overline{A}_{s}=\frac{1}{|\mathcal{B}|}\sum_{b\in\mathcal{B}} A_{b,s},
  \qquad s\in\{\text{overall, action}\},
  \label{eq:sup-general-average}
\end{equation}
where $\mathcal{B}$ denotes the set of four general video-understanding benchmarks ($|\mathcal{B}|=4$). 
Means and gains are computed from unrounded benchmark scores and rounded to one decimal place for display; baselines with unavailable benchmark scores are omitted from the four-benchmark average.
\begin{table}[t]
\centering
\caption{Human inspection of ActionTree-97k annotations across 1,000 randomly sampled examples.}
\label{tab:sup-human-inspection}
\small
\setlength{\tabcolsep}{5pt}
\renewcommand{\arraystretch}{0.96}
\begin{tabular*}{\columnwidth}{@{\extracolsep{\fill}}lrrr@{}}
\toprule
\textbf{Criterion} & \textbf{Inspected} & \textbf{Passed} & \textbf{Rate} \\
\midrule
Visible observations & 1,000 & 972 & 97.2\% \\
Supported hierarchy & 1,000 & 958 & 95.8\% \\
Correct action label & 1,000 & 982 & 98.2\% \\
Faithful uncertainty & 1,000 & 948 & 94.8\% \\
\midrule
Overall acceptable & 1,000 & 936 & 93.6\% \\
\bottomrule
\end{tabular*}
\end{table}

\subsection{Inference Settings and Answer Parsing}
Across all evaluations, inference uses a single sampled response per prompt with near-greedy decoding (temperature $0.01$, top-$p$ $0.001$, and random seed $42$). 
For OVAR benchmarks, videos are sampled at 2~FPS with at most 16 frames and a generation limit of 512 tokens. 
For general video understanding, generation is capped at 1,024 tokens; videos in MVBench, Video-MME, and TempCompass are sampled at 2~FPS with a maximum of 32 frames per video, whereas SEED-Bench directly uses the eight benchmark-supplied input frames.

For OVAR, we extract the predicted action from the model output and compare the normalized label with the ground-truth action. 
For general multiple-choice benchmarks, the parser extracts the option letter (\texttt{A}--\texttt{E}) from the \texttt{<answer>...</answer>} tag, or falls back to terminal option patterns (e.g., \texttt{Answer: X}) and normalized option-text matching when tags are absent.


\section{Additional Analyses and Ablations}
\label{sec:sup-quantitative}

\subsection{Human Inspection of Action-Tree Annotations}
\label{sec:sup-human-inspection}

To complement the automatic verification pipeline used during dataset construction, we manually inspected a random sample of 1,000 ActionTree-97k annotations alongside their corresponding videos. 
The inspection evaluated four criteria: (1) whether leaf observations are directly visible in the sampled frames (\emph{Visible observations}); (2) whether leaves support their parent branches and branches jointly explain the root activity (\emph{Supported hierarchy}); (3) whether the retained action label accurately matches the depicted activity (\emph{Correct action label}); and (4) whether unconfirmable details are properly isolated from observed evidence (\emph{Faithful uncertainty}). 
An annotation was judged overall acceptable only when its hierarchical description was both structurally sound and faithful to the visible video content. 

As shown in Table~\ref{tab:sup-human-inspection}, pass rates across individual criteria range from 94.8\% to 98.2\%, and 936 of the 1,000 inspected annotations (93.6\%) were rated overall acceptable, confirming the high reliability of the verified action trees. 

\textbf{Failure Analysis:} We carefully categorized the remaining 6.4\% of unacceptable cases. The majority of these flaws resulted from extreme motion blur, rapid occlusions of micro-actions (e.g., a hand fleetingly obscuring a small object), or action stages falling outside the sampled 16-frame window. In such edge cases, the verifier occasionally hallucinated visual evidence to bridge temporal gaps, indicating that our strict 16-frame temporal boundary is the primary bottleneck rather than the reasoning capability of the annotator.

\subsection{Component Complementarity and Tree Topology Ablations}
\label{sec:sup-path-hierarchy}

\textbf{Complementary roles of TCM and RTL.} 
Table~3 in the main text reveals a striking divergence in how TCM and RTL contribute to VideoTED. On highly challenging fine-grained tasks like SSv2, on-policy exploration suffers from severe reward sparsity: as analyzed in Sec.~5.3 and Fig.~5 of the main paper, group success starts at only 6.55\% in the first 100 steps, and even in the final 100 steps, 57.4\%--61.8\% of the sampled response groups still yield zero reward across all candidates, entirely eliminating the group-relative advantage gradient. RTL effectively rescues this optimization bottleneck by supplying a reliable offline reference, pulling the SSv2 HM from 15.7\% (GRPO baseline) to 18.3\% (RTL only).
Conversely, when positive responses are successfully sampled, RTL alone struggles to isolate \textit{which} specific visual observations caused the success. TCM steps in to provide dynamic, token-level credit assignment via the self-teacher, driving the General Avg.\ from 62.8\% (RTL only) to 65.2\% (Full). Their integration confirms that RTL guarantees a learning lower bound under sparse success, while TCM raises the performance ceiling.

\begin{table}[t]
\centering
\caption{Effect of hierarchical tree structures (\%). All variants share the exact same textual nodes and labels, differing only in their structural formats and internal connections.}
\label{tab:sup-shuffled-hierarchy}
\small
\setlength{\tabcolsep}{4pt}
\renewcommand{\arraystretch}{0.96}
\begin{tabular*}{\columnwidth}{@{\extracolsep{\fill}}lccc@{}}
\toprule
\textbf{Structural Format} & \textbf{General Avg.} & \textbf{Action Avg.} & \textbf{SSv2 HM} \\
\midrule
w/o Tree Structure (Flat) & 61.1 & 64.5 & 12.4 \\
Shuffled Hierarchy & 63.7 & 70.1 & 17.5 \\
\textbf{Full VideoTED} & \textbf{65.2} & \textbf{71.7} & \textbf{18.4} \\
\bottomrule
\end{tabular*}
\end{table}

\textbf{Does the model learn topology or just formatting?} 
To verify that our performance gains stem from genuine evidence-supporting topology rather than merely the structural JSON format, we introduce a \textit{Shuffled Hierarchy} baseline in Table~\ref{tab:sup-shuffled-hierarchy}. In this setup, we retain the exact same textual nodes as the full model but randomly permute the parent-child connections between leaf observations and sub-action branches. Although the shuffled variant improves over the flat text baseline (63.7\% vs.\ 61.1\% General Avg.), it still consistently trails the full topology (65.2\% General Avg.\ and 18.4\% SSv2 HM), confirming that VideoTED learns to reason over valid semantic dependencies rather than superficially mimicking indentation templates.

\begin{figure*}
    \centering
    \includegraphics[width=\textwidth]{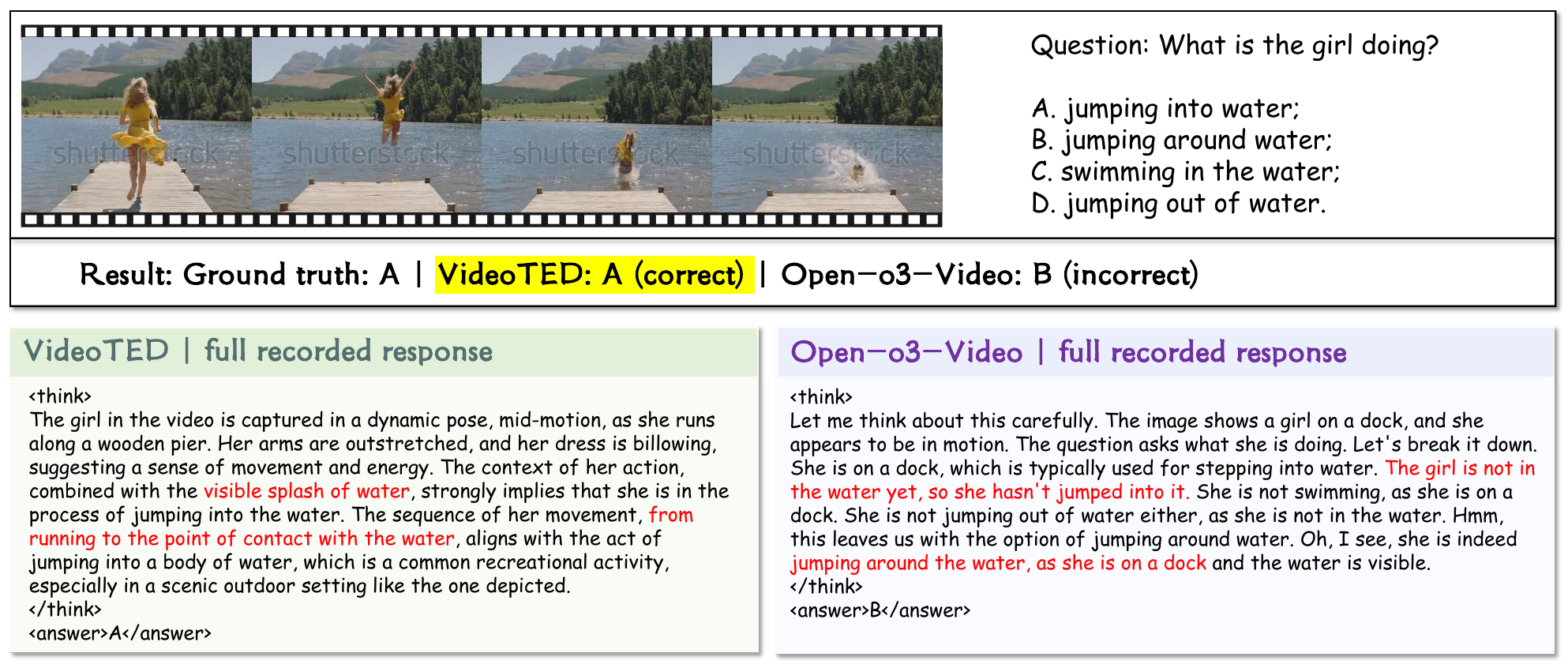}
    \caption{Action-direction example from TempCompass. VideoTED follows
    the visible progression from the pier to water entry and predicts
    A (\emph{jumping into water}); Open-o3-Video predicts B
    (\emph{jumping around water}). The four displayed frames are
    selected from the recorded evaluation input.}
    \label{fig:sup-case-water}
\end{figure*}

\subsection{Sensitivity to Training Coefficients}
\label{sec:sup-sensitivity}

VideoTED integrates Tree-Conditioned Credit Modulation (TCM) and Reference Trajectory Learning (RTL). 
We evaluate the model's sensitivity to three core coefficients: the initial TCM mixing coefficient $\lambda_0$, the RTL loss coefficient $\eta$ , and the level-aware weighting multipliers $\gamma_{jit} \in \{\gamma_{\mathrm{leaf}}, \gamma_{\mathrm{branch}}, \gamma_{\mathrm{root}}\}$.
We vary one configuration at a time while holding the others fixed at their adopted defaults ($\lambda_0=0.5$, $\eta=0.1$, and $\gamma_{jit}=\{1, 2, 4\}$). 
All variants use the Qwen2.5-VL-7B-Instruct backbone and the full ActionTree-97k dataset. Setting $\lambda_0=0$ or $\eta=0$ disables TCM or RTL, respectively. 

\begin{table}[t]
\centering
\caption{Sensitivity to the TCM coefficient $\lambda_0$, RTL coefficient $\eta$, and TCM level weighting $\gamma_{jit}$ (\%). Bold values indicate the adopted default configuration.}
\label{tab:sup-sensitivity}
\small
\setlength{\tabcolsep}{2pt}
\renewcommand{\arraystretch}{0.96}
\begin{tabular*}{\columnwidth}{@{\extracolsep{\fill}}lccc@{}}
\toprule
\textbf{Hyperparameter} & \textbf{General Avg.} & \textbf{Action Avg.} & \textbf{SSv2 HM} \\
\midrule
\rowcolor[RGB]{232,237,247}
\multicolumn{4}{@{}l}{\emph{Initial TCM coefficient $\lambda_0$ (with $\eta=0.1, \gamma=\{1,2,4\}$)}} \\
$\lambda_0 = 0.0$ (w/o TCM) & 62.8 & 69.2 & 18.3 \\
$\lambda_0 = 0.25$ & 64.4 & 70.8 & \textbf{18.4} \\
$\mathbf{\lambda_0 = 0.50}$ & \textbf{65.2} & \textbf{71.7} & \textbf{18.4} \\
$\lambda_0 = 0.75$ & 64.6 & 71.1 & 18.1 \\
\midrule
\rowcolor[RGB]{232,237,247}
\multicolumn{4}{@{}l}{\emph{RTL reference coefficient $\eta$ (with $\lambda_0=0.5, \gamma=\{1,2,4\}$)}} \\
$\eta = 0.0$ (w/o RTL) & 62.8 & 68.4 & 17.2 \\
$\eta = 0.05$ & 64.3 & 70.6 & 18.1 \\
$\mathbf{\eta = 0.10}$ & \textbf{65.2} & \textbf{71.7} & \textbf{18.4} \\
$\eta = 0.20$ & 64.5 & 70.9 & 18.2 \\
\midrule
\rowcolor[RGB]{232,237,247}
\multicolumn{4}{@{}l}{\emph{ TCM Level weighting $\gamma_{\mathrm{leaf}} : \gamma_{\mathrm{branch}} : \gamma_{\mathrm{root}}$ (with $\lambda_0=0.5,\eta=0.1$)}} \\
$1.0 : 1.0 : 1.0$ (Uniform) & 64.1 & 70.3 & 18.0 \\
$\mathbf{1.0 : 2.0 : 4.0}$ (Default)& \textbf{65.2} & \textbf{71.7} & \textbf{18.4} \\
$1.0 : 4.0 : 8.0$ (Aggressive) & 64.7 & 70.9 & 18.1 \\
\bottomrule
\end{tabular*}
\end{table}

 \textbf{Analysis of coefficient sweeps.} 
As shown in Table~\ref{tab:sup-sensitivity}, completely removing either component ($\lambda_0=0$ or $\eta=0$) causes a clear drop across all metrics. 
Across positive settings, VideoTED exhibits stable performance over both $\lambda_0 \in [0.25, 0.75]$ and $\eta \in [0.05, 0.20]$. 
Increasing $\lambda_0$ from $0.25$ to $0.50$ strengthens token-level credit assignment on intermediate visual observations, whereas an overly large weight ($\lambda_0=0.75$) slightly over-amplifies early teacher preferences. 
Similarly, $\eta=0.10$ provides sufficient positive supervision when sampled groups fail to achieve reward variance, whereas doubling the weight to $\eta=0.20$ places excessive emphasis on offline reference imitation at the slight expense of on-policy exploration. 
For level-aware weighting, the uniform variant ($\gamma = 1:1:1$) treats verbose low-level visual descriptions identically to the critical activity root, yielding sub-optimal understanding (64.1\% General Avg.). Conversely, an overly aggressive ratio ($\gamma = 1:4:8$) suppresses the learning of visual grounding leaves. The adopted $1:2:4$ ratio strikes the best balance, successfully enforcing the hierarchical concentration of evidence.


\begin{figure*}[t]
    \centering
    \includegraphics[width=\textwidth]{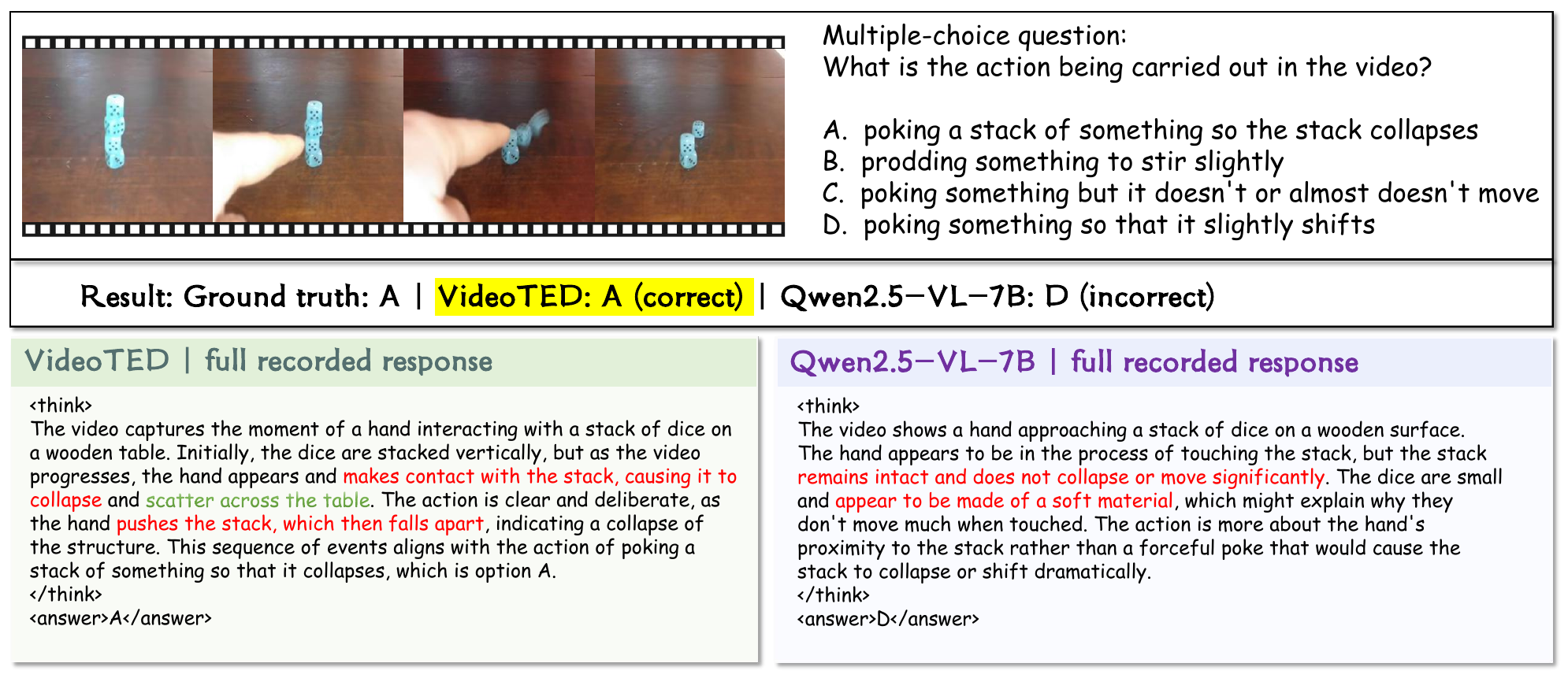}
    \caption{Object-state example from the SEED-Bench video subset. VideoTED predicts A (\emph{poking a stack of something so the stack collapses}), matching the ground truth; Qwen2.5-VL-7B predicts D (\emph{poking something so that it slightly shifts}). The four displayed frames are selected from the benchmark's eight supplied input frames.}
    \label{fig:sup-case-dice}
\end{figure*}

\begin{figure*}[t]
    \centering
    \includegraphics[width=\textwidth]{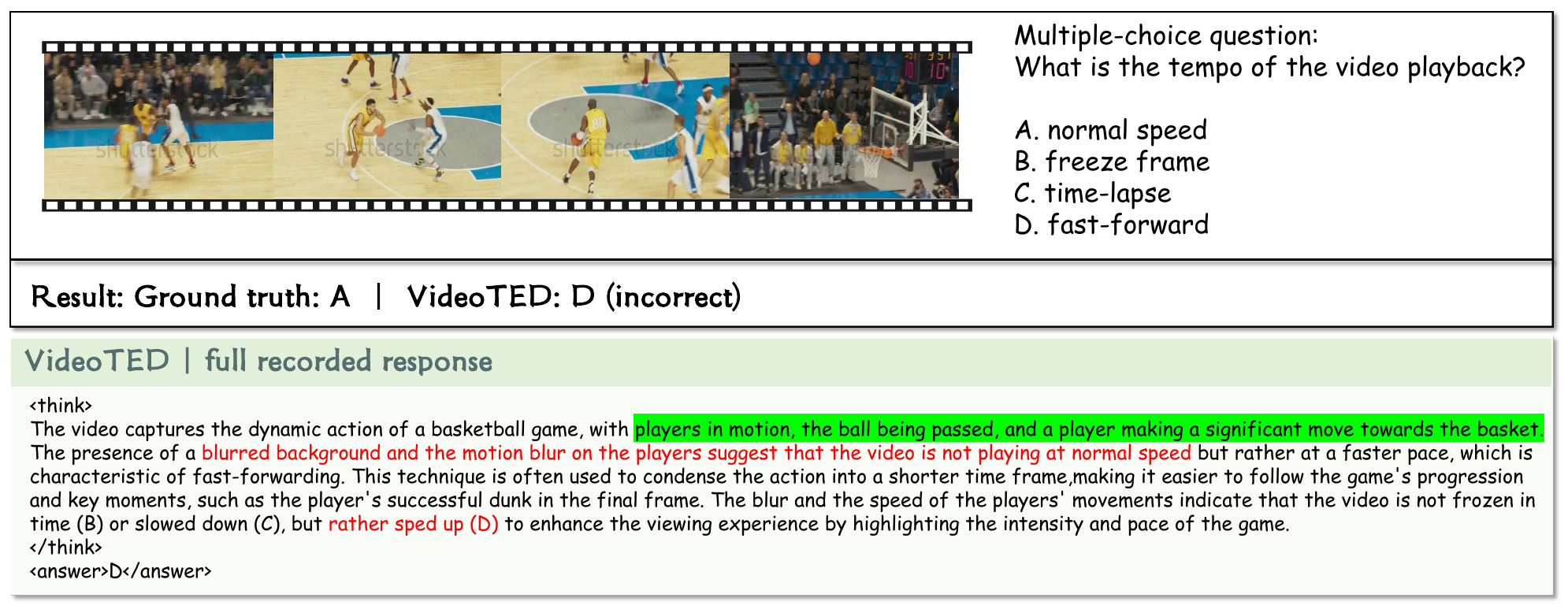}
    \caption{Playback-speed failure case from TempCompass. The ground-truth answer is A (\emph{normal speed}), while VideoTED predicts D (\emph{fast-forward}). VideoTED accurately recognizes the basketball actions but misinterprets rapid player motion and motion blur as accelerated playback. The four displayed frames are selected from the recorded evaluation input.}
    \label{fig:sup-case-fail-tempcompass}
\end{figure*}

\section{Qualitative Case Studies}
\label{sec:sup-qualitative}

To qualitatively illustrate how VideoTED reasons over temporal progression and object state changes, we present two representative success cases from TempCompass and SEED-Bench alongside one failure case. These examples are intended as qualitative analyses of reasoning behavior rather than separate accuracy measurements.

\textbf{Resolving action direction from later frames.}
Figure~\ref{fig:sup-case-water} presents a TempCompass Action Direction question that distinguishes jumping \emph{into} water from jumping \emph{around} it. The selected frames show the girl running along the pier, airborne over the water, entering it, and producing a visible splash. VideoTED links this temporal progression to the water-entry option and predicts A, matching the ground truth. In contrast, Open-o3-Video overlooks the later frames, concludes that the girl has not yet entered the water, and selects B. This contrast highlights how tracking the action endpoint is decisive for direction-sensitive queries.

\textbf{Recognizing a change in object configuration.}
Figure~\ref{fig:sup-case-dice} shows a SEED-Bench Action Recognition question distinguishing a collapsing stack from a slight shift. The supplied frames begin with a vertical stack of dice, followed by a finger approaching, the upper dice toppling, and the dice separated in the final frame. VideoTED captures this state transition and selects A, matching the ground truth, whereas Qwen2.5-VL-7B claims the stack remains intact and selects D. Notably, VideoTED's reasoning states that the dice ``scatter across the table,'' which is slightly stronger than the sampled frames alone establish. Thus, a correct final prediction does not guarantee that every descriptive detail in the reasoning trace is strictly grounded.

\textbf{Failure case: distinguishing fast motion from fast playback.}
Figure~\ref{fig:sup-case-fail-tempcompass} presents a TempCompass Speed question featuring a fast-paced basketball sequence played at normal speed. While VideoTED accurately identifies the core on-court actions---including players in motion, passing, and driving toward the basket---it treats the blurred background and motion blur on the players as evidence of accelerated playback, predicting D (\emph{fast-forward}) instead of the ground-truth A (\emph{normal speed}). This failure illustrates a confusion between rapid in-scene motion and sped-up playback under fixed-rate sampling.


\section{Limitations and Future Work}
\label{sec:sup-limitations}

ActionTree-97k relies on model-generated evidence trees constructed from sampled video frames. Although the automatic verification procedure strictly checks the prescribed visual support, hierarchical consistency, and label conditions, automatically constructed annotations are not entirely equivalent to exhaustive human-verified descriptions; residual noise in a reference tree can still affect both the privileged teacher context and the reference response. In addition, the fixed-rate frame sampler may miss intermediate states during brief contacts, rapid action transitions, or severe motion blur, leading to occasional ambiguities in fine-grained action or speed-related judgments.

Future work can build upon these observations along several directions. First, incorporating adaptive sampling or keyframe selection around rapid transitions could capture brief contacts and subtle state changes more completely while keeping computational overhead manageable. Second, coupling hierarchical tree nodes with explicit timestamps or spatial grounding offers a promising avenue to further improve evidence traceability and fine-grained perception. Finally, extending VideoTED's tree-guided evidence distillation framework to longer-horizon videos with multiple interleaved events remains an interesting direction for future exploration.

\end{document}